\documentclass[11pt,a4paper]{article}

\usepackage[utf8]{inputenc}
\usepackage[T1]{fontenc}
\usepackage{lmodern}

\usepackage[margin=1in]{geometry}
\usepackage{setspace}
\usepackage{amsmath,amssymb,amsthm}

\usepackage{graphicx}
\graphicspath{{figures/}}
\usepackage{booktabs}
\usepackage{float}
\usepackage[section]{placeins}

\usepackage[numbers,square]{natbib}
\usepackage{hyperref}
\hypersetup{
  colorlinks=true,
  linkcolor=blue!60!black,
  citecolor=blue!60!black,
  urlcolor=blue!60!black,
  breaklinks=true
}
\usepackage{cleveref}

\usepackage{xcolor}


\begin{document}

\title{Whether, Not Which: Mechanistic Interpretability Reveals\\
Dissociable Affect Reception and Emotion Categorization in LLMs}

\author{Michael Keeman\\
Keido Labs, Liverpool, UK\\
\texttt{michael@keidolabs.com}}

\date{}

\maketitle

% ============================================================================
% ABSTRACT
% ============================================================================

\begin{abstract}
Large language models appear to develop internal representations of emotion---``emotion circuits,'' ``emotion neurons,'' and structured emotional manifolds have been reported across multiple model families.
But every study making these claims uses stimuli where emotional content is signalled by explicit emotion keywords, leaving a fundamental question unanswered: do these circuits detect genuine emotional meaning, or do they detect the word ``devastated''?
We present the first clinical validity test of emotion circuit claims using mechanistic interpretability methods grounded in clinical psychology.
Rather than crowdsourced text, we use clinical vignettes that evoke specific emotions through situational and behavioural cues alone, with emotion keywords systematically removed---stimuli grounded in established clinical psychology methodology.
Across six models (Llama-3.2-1B, Llama-3-8B, Gemma-2-9B; base and instruct variants), we apply four convergent mechanistic interpretability methods---linear probing, causal activation patching, knockout experiments, and representational geometry---and discover two dissociable emotion processing mechanisms.
\emph{Affect reception}---the detection of emotionally significant content---operates with near-perfect accuracy (AUROC 1.000), a result consistent with early-layer saturation rather than overfitting, and replicates across all six models.
\emph{Emotion categorization}---the mapping of detected affect to specific emotion labels---is partially keyword-dependent, dropping 1--7\% without keywords and improving with scale.
Cross-set causal activation patching confirms that keyword-rich and keyword-free stimuli share representational space, with patched activations transferring an affective salience signal rather than emotion-category identity.
These findings falsify the keyword-spotting hypothesis, establish a novel mechanistic dissociation, and introduce clinical stimulus methodology as a rigorous external standard for testing emotion processing claims in large language models.
These results have direct implications for AI safety evaluation, where emotion-processing claims increasingly inform alignment and deployment decisions.
All stimuli, code, and data are released for replication.

\end{abstract}

% ============================================================================
% 1. INTRODUCTION
% ============================================================================
\section{Introduction}
\label{sec:introduction}

A kitchen table set for two, as usual.
One plate untouched, the coffee cold.
Across from her seat, his photo and a small urn\ldots

A human reader recognizes grief instantly---not from any emotion word, but from the situation itself.
A standard sentiment analyzer does not: zero emotion keywords, one incidental valence hit (``cold''---referring to coffee).
Emotionally invisible.

The question this paper asks is whether a large language model sees what the human sees or what the sentiment analyzer sees.

Recent mechanistic interpretability studies paint an increasingly detailed picture of emotion processing inside LLMs.
\citet{takMechanisticInterpretabilityEmotion2025} demonstrate that emotion information consolidates in mid-layer activations and can be causally transferred via activation patching.
\citet{leeLargeLanguageModels2025} identify neurons that fire selectively on emotion-laden text.
\citet{reichmanEmotionsWhereArt2026} map low-dimensional emotional manifolds stable across datasets and languages.
\citet{wangLLMsFeelEmotion2025} show that prompted emotion circuits can steer model outputs.
Collectively, these findings suggest that LLMs develop structured internal representations of emotion---not surface-level sentiment detection, but something closer to a functional encoding of emotional meaning.

These findings represent the first serious steps toward understanding emotional processing in LLMs.
They also leave open the question this study aims to answer: are these circuits detecting emotional meaning, or detecting the keywords that co-occur with it?

Every existing study uses stimuli where emotional content is signaled by explicit emotion vocabulary.
The datasets---crowd-enVENT~\citep{troianoDimensionalModelingEmotions2023}, GoEmotions~\citep{demszkyGoEmotionsDatasetFineGrained2020}, DailyDialog~\citep{liDailyDialogManuallyLabelled2017}, or synthetically generated prompts---contain phrases like ``I was devastated,'' ``she was furious,'' or ``the news filled me with joy.''
When a linear probe decodes ``sadness'' from the residual stream of a model processing ``I was devastated by the loss,'' it faces an ambiguity that the methodology cannot resolve: is the probe detecting the emotional meaning of the situation, or is it detecting the lexical features of the word ``devastated''?

The concern is not new.
\citet{leeLargeLanguageModels2025} note in their discussion that identified emotion neurons may fire on cohyponyms or syntactic frames rather than emotional content per se.
But nobody has run the controlled test.
This paper does.

We present six LLMs with clinically designed emotional vignettes---stimuli that evoke specific emotions at controlled intensities through situational and behavioral cues, with emotion keywords systematically removed.
A grief vignette describes that empty kitchen table.
A rage vignette depicts scattered papers across a conference table after a falsified report surfaces.
No emotion words.
No sentiment phrases.
No internal state descriptions.
Each emotion is crossed with three distinct topic domains, controlling for the possibility that the model encodes topic rather than emotion.

We compare internal processing of these clinical vignettes to standard keyword-rich stimuli across six models (Llama-3.2-1B, Llama-3-8B, Gemma-2-9B; base and instruct variants), using four convergent methodological streams: linear probing (correlational), activation patching (causal), knockout experiments (necessity), and representational geometry (structural).
This convergence within a single study is itself a methodological contribution---prior work uses one or two of these methods, never all four on the same stimulus contrast.

The results reveal a more nuanced picture than either side of the current debate anticipates.
We discover two dissociable mechanisms:

\begin{enumerate}
  \item \textbf{Affect reception}---the detection of emotional significance---is keyword-independent.
  Binary probes (emotional vs.\ neutral) achieve AUROC 1.000 on clinical vignettes across all six models, reaching ceiling in early layers.
  The models know that something emotionally significant is happening, from pure situational context alone.

  \item \textbf{Emotion categorization}---the mapping of detected affect to specific emotion categories---is partially keyword-dependent.
  Eight-class probe AUROC drops 1--7\% on clinical vignettes relative to keyword-rich stimuli, with larger models showing smaller drops (1.1--1.9\% at 8B/9B vs.\ 4.6--6.7\% at 1B).
\end{enumerate}

We did not expect the first finding.
Near-perfect binary detection without any emotion keywords---across all six models, two families, base and instruct---was surprising enough that we ran dedicated confirmation experiments to rule out artifacts.
High-complexity neutral narratives matched on sensory detail and sentence structure scored 0.04 where emotional vignettes scored 0.999 (\cref{app:setc}).
Zero-shot controls confirmed the signal is intrinsic to target processing, not inherited from few-shot context.
The binary probes can be trusted: what they detect is genuine emotional significance, not narrative complexity or prompt leakage.

Cross-set activation patching provides causal confirmation: patching keyword-rich activations into keyword-free forward passes transfers an \emph{affective salience signal} that boosts categorization accuracy regardless of the source emotion's identity.
This is direct evidence that affect reception and emotion categorization operate through separable causal pathways.

This paper makes four contributions.
First, it provides the first clinical validity test of emotion circuit claims in LLMs, using keyword-free stimuli grounded in clinical methodology.
Second, it discovers the affect reception / emotion categorization dissociation---a mechanistic distinction not previously reported.
Third, it provides convergent evidence (correlational, causal, necessity, and geometric) within a single study, across six models from two families.
Fourth, it releases everything---96 clinical vignettes, extraction pipeline, analysis scripts, result data---as an open-source resource for the community.

\begin{figure}[!htbp]
  \centering
  \includegraphics[width=\textwidth]{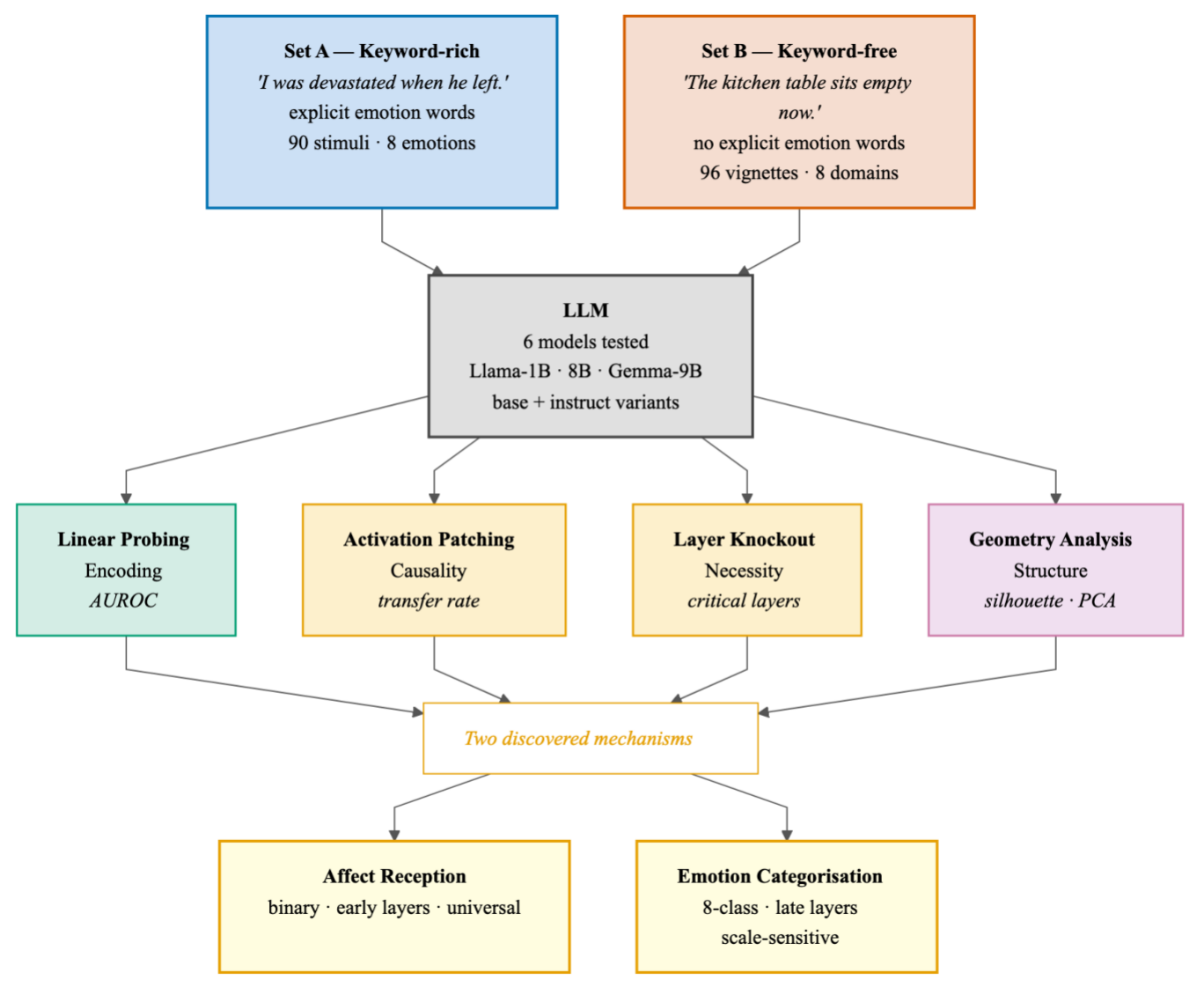}
  \caption{Experimental design overview.
  Both keyword-rich stimuli (Set~A; explicit emotion vocabulary) and clinical vignettes (Set~B; situational cues, no emotion words) are processed by the same six LLMs.
  Four convergent analysis methods---linear probing, activation patching, knockout experiments, and representational geometry---probe internal activations at every layer, yielding evidence for two dissociable mechanisms: \emph{affect reception} (binary, early-layer, universal) and \emph{emotion categorisation} (8-class, mid-to-late layer, scale-sensitive).}
  \label{fig:schematic}
\end{figure}

% ============================================================================
% 2. RELATED WORK
% ============================================================================
\section{Related Work}
\label{sec:related-work}

\subsection{Emotion Circuits in LLMs}
\label{sec:emotion-circuits}

The mechanistic study of emotion processing in LLMs is recent but growing rapidly.
\citet{takMechanisticInterpretabilityEmotion2025} provide the most comprehensive treatment, demonstrating that emotion is linearly decodable from mid-layer activations, causally transferable via activation patching, and steerable using emotion direction vectors.
Their study spans multiple model families and sizes, establishing mid-layer consolidation as a robust phenomenon.
We treat their work as our primary replication target: we first reproduce their methodology with their stimulus type, then extend it with clinical stimuli that test whether their findings survive keyword removal.

\citet{leeLargeLanguageModels2025} ask ``Do LLMs Have Emotion Neurons?'' and identify neurons that fire selectively on emotional text.
Their discussion section is, in effect, our research question---they note that identified neurons may respond to cohyponyms or syntactic frames rather than emotional content, but do not test this possibility.
We operationalize their concern as an experimental contrast.

\citet{reichmanEmotionsWhereArt2026} map an ``emotional manifold''---a low-dimensional subspace of model activations organized by emotion---and demonstrate that learned intervention modules can steer emotion perception along these directions.
Their finding that emotion directions generalize across datasets and languages is suggestive of genuine emotional representation, but their stimuli share the same keyword confound.

\citet{zhangDecodingEmotionDeep2025} examine emotion encoding depth in a systematic layer-wise probing study, showing that emotional signals form clusters in representation space that are sharper internally than what zero-shot output classification reveals.
Their finding that emotional information persists over hundreds of subsequent tokens is consistent with the affect reception mechanism we report here.

Additional evidence for structured internal emotion geometry comes from work on neural emotion decoding~\citep{vosDecodingNeuralEmotion2025}, circuit-level emotion manipulation~\citep{wangLLMsFeelEmotion2025}, and emotion-sensitive neurons in audio-language models~\citep{zhaoDiscoveringCausallyValidating2026}, extending the picture across modalities and methodologies.

Beyond emotion specifically, \citet{chenPersonaVectorsMonitoring2025a} identify linear directions in LLM activation space underlying character traits including sycophancy, hallucination propensity, and ethical alignment---\emph{persona vectors}.
Their finding that these directions predict and track personality shifts after finetuning, and can be used to monitor trait fluctuations at deployment time, establishes a broader paradigm: psychological properties of language models are linearly represented in activation space and causally relevant to behavior.
Our work on emotion processing operates within this paradigm, extending it with a clinical validity criterion that prior work---on both emotion and character---has not applied.

At the theoretical level, the Persona Selection Model~\citep{PersonaSelectionModel} proposes that LLMs learn to simulate diverse personas during pre-training, with emotional representations among the character-consistent features installed across the training distribution.
This framework generates two testable predictions relevant to our study: that emotional encoding should be a pre-training phenomenon---observable in base models before any alignment training---and that psychological methods designed for assessing human emotional processing should transfer productively to AI internal representations.
We test both predictions directly (\cref{sec:tuning-interpretation,sec:psm}).

\subsection{The Keyword Confound as a Clinical Validity Problem}
\label{sec:keyword-confound}

All existing studies use stimuli drawn from crowd-enVENT~\citep{troianoDimensionalModelingEmotions2023}, GoEmotions~\citep{demszkyGoEmotionsDatasetFineGrained2020}, DailyDialog~\citep{liDailyDialogManuallyLabelled2017}, or synthetic prompts containing explicit emotion vocabulary.
This is not a minor methodological quibble---it is a clinical validity problem.
In psychology, a measure of emotional processing must demonstrate that it responds to emotional \emph{meaning}, not to the surface markers that correlate with emotional meaning.
A depression screener that only detects patients who say ``I am depressed'' has no clinical validity.
The same standard applies to internal-representation claims about LLMs: can a mechanistic finding survive removal of lexical markers while preserving ecological emotional meaning?
If not, the claim describes keyword detection, not emotion processing.

The confound is acknowledged but untested.
What is needed is straightforward: stimuli that carry emotional meaning without carrying emotion vocabulary.
If probes still decode emotion from these, the keyword-spotting explanation fails.
If they do not, the existing claims need revision.
Separately, \citet{schaefferAreEmergentAbilities2023} demonstrate that metric choice alone can create illusory discontinuities in the emergent abilities literature---a cautionary example of how measurement artifacts masquerade as genuine capabilities.
We follow their recommendation to use continuous metrics (AUROC rather than accuracy) throughout.

\subsection{Clinical Stimulus Methodology}
\label{sec:clinical-stimulus}

Clinical psychology has spent decades developing rigorous methodologies for creating text-based stimuli that evoke specific emotional responses at controlled intensities.
Validated verbal emotion vignettes~\citep{wingenbachDevelopmentValidationVerbal2019} demonstrate that carefully constructed situational narratives reliably elicit targeted emotions through context rather than explicit labeling, with hit rates above 80\% and intensity ratings in the 65--90\% range across anger, grief, and fear conditions.
Crisis vignettes from the LEAP tradition~\citep{harrisInterventionCrisisUnderstanding2025} show that factorial vignette designs---varying one situational factor at a time---isolate emotional content from confounding contextual variables while maintaining ecological validity.
PTSD script-driven imagery protocols~\citep{langFearBehaviorFear1983,pitmanPsychophysiologicAssessmentPosttraumatic1987} establish that first-person situational narratives with no explicit emotional labeling reliably produce differential physiological and psychological responses across emotion categories.
Grief narrative design has its own methodological literature, with established guidelines for constructing bereavement stimuli that elicit genuine emotional salience through situational description rather than internal state disclosure~\citep{hossainDesigningGriefPACER2025}.
Vignette factorial methodology more broadly has been validated as a tool for isolating psychological constructs in controlled experimental designs~\citep{atzmullerExperimentalVignetteStudies2010}.

To our knowledge, clinical stimulus methodology has not been applied to mechanistic analysis of internal model states.
The tools exist in clinical psychology.
The gap is in mechanistic interpretability, where nobody has used them.

\subsection{Affect Reception in Affective Neuroscience}
\label{sec:affect-reception-neuro}

The term ``affect reception'' draws on established neuroscience of receptive emotional processing.
The human brain processes emotionally charged input through rapid, parallel circuits: the ``low road'' subcortical pathway (thalamus $\to$ amygdala, ${\sim}12$\,ms) provides crude valence and arousal tagging before the ``high road'' cortical pathway (${\sim}40$\,ms) enables categorical emotion identification~\citep{LeDoux1996Emotional,ledouxRethinkingEmotionalBrain2012}.
This produces a temporal dissociation: organisms detect ``something emotionally significant is happening'' before they identify what specific emotion is present.
Affective neuroscience further distinguishes \emph{receptive} emotional processing---the automatic detection and encoding of emotional significance in input---from \emph{expressive} processing---the generation of emotional behavior and response---as separable systems with distinct neural substrates, timing, and modulating conditions~\citep{adolphsRecognizingEmotionFacial2002,fusar-poliFunctionalAtlasEmotional2009}.
Early ERP components (P1/N170, ${\sim}100$--$200$\,ms) index receptive salience detection independent of any motor or expressive response; categorical identification requires later cortical processing (${\sim}250$--$400$\,ms)~\citep{schuppSelectiveVisualAttention2007}.

We are not claiming that LLMs have amygdalas or that transformer layers map onto subcortical pathways.
We use this framework as a \emph{structural analogy} for naming the dissociation we observe: early, robust, keyword-independent detection of affective significance (affect reception) followed by later, more fragile, partially keyword-dependent categorical discrimination (emotion categorization).
The analogy is functional, not mechanistic.
But it is not arbitrary---the same information-theoretic logic applies.
Detecting that something is emotionally significant is a simpler computation than determining which specific emotion is present, and the simpler computation should emerge first, more robustly, and with less dependence on surface features.
That is exactly what we find.

% ============================================================================
% 3. STIMULI AND MODELS
% ============================================================================
\section{Stimuli and Models}
\label{sec:stimuli-models}

\subsection{Stimulus Set~A --- Keyword-Rich (Replication Baseline)}
\label{sec:set-a}

Set~A comprises 80 emotional stimuli (10 per emotion $\times$ 8 Plutchik primary emotions~\citep{Plutchik1980General}: joy, trust, fear, surprise, sadness, disgust, anger, anticipation, mapped to their peak intensities: ecstasy, admiration, terror, amazement, grief, loathing, rage, vigilance).
Stimuli are drawn from crowd-enVENT~\citep{troianoDimensionalModelingEmotions2023} and contain explicit emotion vocabulary (``I was devastated,'' ``furious,'' etc.), matching the stimulus type used in~\citet{takMechanisticInterpretabilityEmotion2025} and~\citet{leeLargeLanguageModels2025}.

\subsection{Stimulus Set~B --- Clinical Vignettes (Keyword-Free)}
\label{sec:set-b}

Set~B comprises 96 clinical vignettes (8 emotions $\times$ 3 topic domains $\times$ 4 vignettes per cell) designed by a clinical psychologist (the author).
The keyword control protocol requires: no emotion words, no sentiment phrases, no internal state descriptions.
Emotional content is conveyed exclusively through situational and behavioral cues.
As a preflight validation, all 96 vignettes were run through the keyword-based sentiment baseline (\cref{sec:sentiment-baseline})---the overwhelming majority registered zero emotion keyword hits, confirming that any emotion signal recovered by neural probes is not attributable to residual lexical cues.

The cross-topic design is the primary confound control.
Each emotion spans three distinct topic domains---grief, for instance, is conveyed through bereavement, career destruction, and relationship dissolution scenarios.
If the model encodes \emph{topic} rather than \emph{emotion}, grief-bereavement vignettes should cluster with bereavement-related vignettes of other emotions rather than with grief-career vignettes.
The cross-topic permutation test (\cref{sec:permutation-test}) directly evaluates this.

Ninety-six matched neutral controls (same topic domains, no emotional content) enable the binary detection contrast.

\begin{table}[ht]
  \centering
  \caption{Stimulus set comparison.}
  \label{tab:stimulus-comparison}
  \begin{tabular}{lll}
    \toprule
    Feature & Set~A (Keyword-Rich) & Set~B (Clinical) \\
    \midrule
    N stimuli & 80 & 96 + 96 neutral \\
    Source & crowd-enVENT & Clinical design \\
    Keywords present & Yes & No \\
    Emotions & 8 Plutchik primaries & 8 Plutchik primaries \\
    Topic control & None & 3 domains/emotion \\
    Designer & Crowdsourced & Clinical psychologist \\
    \bottomrule
  \end{tabular}
\end{table}

\subsection{Prompt Format}
\label{sec:prompt-format}

All experiments use the~\citet{takMechanisticInterpretabilityEmotion2025} few-shot classification format.
Two fixed keyword-rich few-shot examples precede the target stimulus.
The model's prediction is extracted at the \texttt{:} token position (the colon following ``Answer'').
The few-shot examples use keyword-rich demonstrations for \emph{both} stimulus sets---a deliberate design choice that tests whether the model can generalize from keyword-rich context to keyword-free targets.

A zero-shot control experiment confirms that the few-shot context contributes no signal to the encoding: Llama-1B instruct achieves identical binary AUROC (1.000) and marginally higher 8-class AUROC (0.938 vs.\ 0.934) without few-shot examples.
The observed encoding is intrinsic to the model's processing of the target text.

\subsection{Keyword-Based Sentiment Baseline}
\label{sec:sentiment-baseline}

To verify that our clinical vignettes resist detection by classical NLP methods, we applied a lexicon-based sentiment analysis pipeline.
Each stimulus was tokenized and matched against curated keyword lists spanning six emotion categories (anger, fear, sadness, joy, disgust, surprise; ${\sim}20$--$25$ terms each) and positive/negative valence dictionaries (${\sim}35$ terms each), modeled on the NRC Emotion Lexicon and VADER frameworks.
Polarity was computed as $(\text{pos} - \text{neg}) / (\text{pos} + \text{neg})$.
Stimuli with zero emotion keyword hits and ${\leq}1$ valence word were classified as ``invisible'' to keyword analysis.

The result: Set~B clinical vignettes are overwhelmingly invisible to keyword-based methods.
This establishes that any emotion discrimination achieved by the neural network probes (AUROC 0.93--0.99 on Set~B) cannot be attributed to surface-level lexical cues available to bag-of-words sentiment classifiers.
Whatever the models are detecting, it is not keywords.

\subsection{Models}
\label{sec:models}

\begin{table}[ht]
  \centering
  \caption{Model specifications.}
  \label{tab:models}
  \begin{tabular}{lrcrl}
    \toprule
    Model & Parameters & Layers & $d_\text{model}$ & Type \\
    \midrule
    Llama-3.2-1B & 1.24B & 16 & 2048 & Base \\
    Llama-3.2-1B-Instruct & 1.24B & 16 & 2048 & Instruct \\
    Llama-3-8B & 8.03B & 32 & 4096 & Base \\
    Llama-3-8B-Instruct & 8.03B & 32 & 4096 & Instruct \\
    Gemma-2-9B & 9.24B & 42 & 3584 & Base \\
    Gemma-2-9B-it & 9.24B & 42 & 3584 & Instruct \\
    \bottomrule
  \end{tabular}
\end{table}

Models span two open-weight families: Llama-3.2-1B and Llama-3-8B from Meta AI's Llama~3 series~\citep{grattafioriLlama3Herd2024}, and Gemma-2-9B from Google DeepMind~\citep{teamGemma2Improving2024}.
Each is tested in base and instruction-tuned variants, yielding six models total.

All experiments run on consumer hardware (MacBook Pro M1 Max, 32\,GB unified memory) using float16 precision.
This is a deliberate reproducibility choice: our complete pipeline requires no GPU cluster access.

% ============================================================================
% 4. REPLICATION
% ============================================================================
\section{Replication: Emotion Encoding in Keyword-Rich Stimuli}
\label{sec:replication}

Before testing whether emotion encoding survives keyword removal, we first confirm that it exists.
Layer-wise linear probes (LogisticRegression, 8-class, 5-fold stratified cross-validation) are trained on three activation types at each layer: residual stream ($\mathbf{h}$), multi-head self-attention output ($\mathbf{a}$), and feed-forward network output ($\mathbf{m}$).

\begin{table}[ht]
  \centering
  \caption{Set~A peak probe AUROC per model (residual stream).}
  \label{tab:set-a-peak}
  \begin{tabular}{lccc}
    \toprule
    Model & Peak AUROC & Peak Layer & Norm.\ Depth \\
    \midrule
    Llama-1B instruct & 0.999 & L11 & 0.75 \\
    Llama-1B base & 1.000 & L12 & 0.75 \\
    Llama-8B instruct & 1.000 & L32 & 1.00 \\
    Llama-8B base & 1.000 & L18 & 0.56 \\
    Gemma-9B instruct & 1.000 & L22 & 0.52 \\
    Gemma-9B base & 1.000 & L33 & 0.79 \\
    \bottomrule
  \end{tabular}
\end{table}

All six models achieve AUROC ${\geq}\,0.998$ at peak layers across all activation types.
The MHSA component peaks at mid-layer (0.43--0.75 normalized depth), while the FFN component peaks later (0.43--0.97).
This pattern---MHSA consolidation preceding FFN consolidation---suggests that the ``mid-layer'' finding reported by~\citet{takMechanisticInterpretabilityEmotion2025} may reflect a MHSA-dominant signal rather than a uniform mid-layer phenomenon.
Replication succeeds.

\begin{figure}[!htbp]
  \centering
  \includegraphics[width=0.9\textwidth]{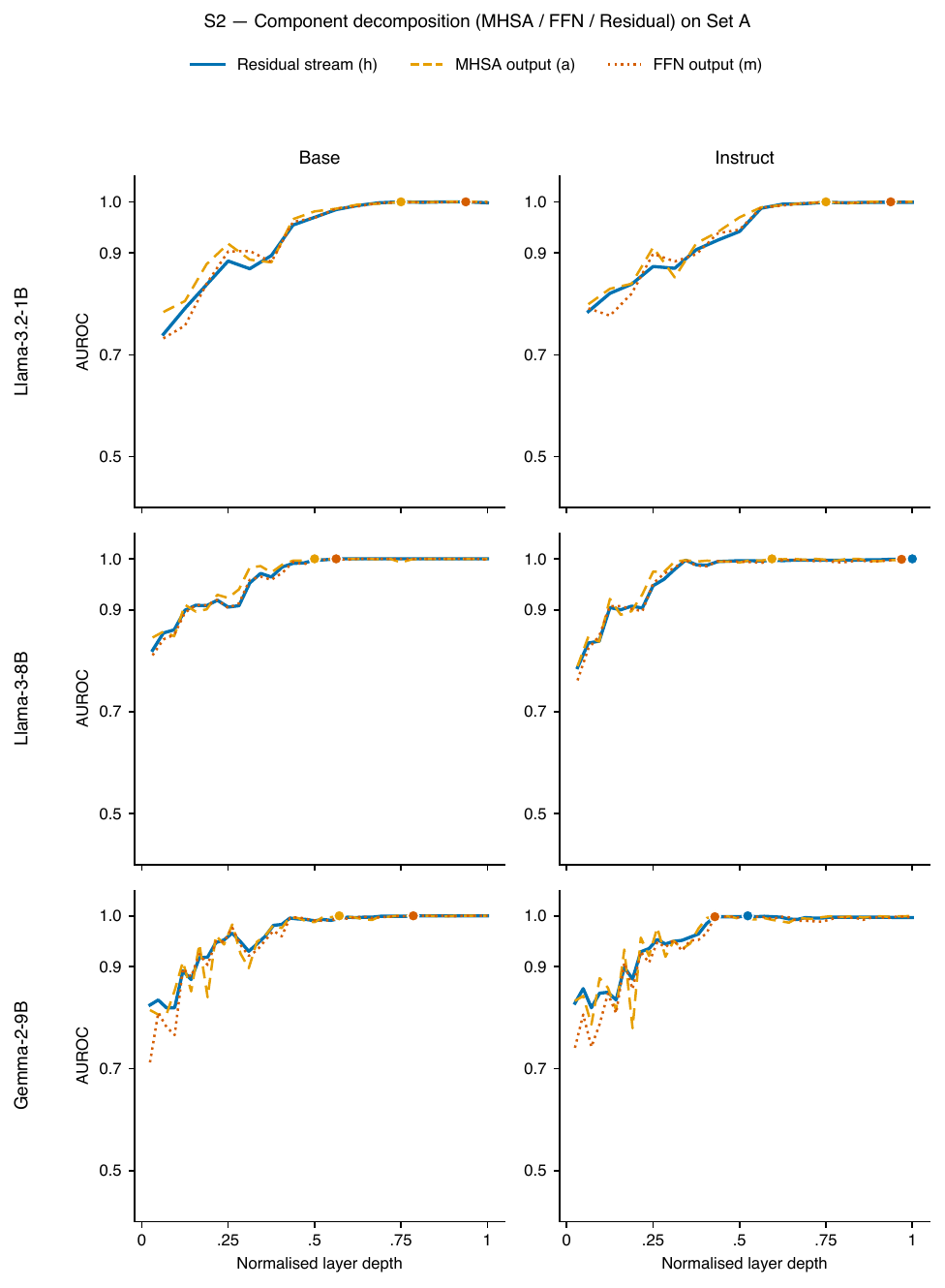}
  \caption{Component decomposition of emotion encoding on Set~A (keyword-rich stimuli).
  Layer-wise 8-class AUROC for three activation types---residual stream ($\mathbf{h}$, solid blue), MHSA output ($\mathbf{a}$, dashed orange), and FFN output ($\mathbf{m}$, dotted red)---across all six models.
  The MHSA component reaches peak AUROC at mid-layer (0.43--0.75 normalized depth), while the FFN component peaks later (0.43--0.97).
  All three components converge to ceiling by the final layers.
  The earlier MHSA peak suggests that multi-head self-attention consolidates emotion information before feed-forward layers process it---a temporal dissociation within the standard transformer sublayer sequence.}
  \label{fig:components}
\end{figure}

% ============================================================================
% 5. THE CLINICAL VALIDITY TEST
% ============================================================================
\section{The Clinical Validity Test}
\label{sec:clinical-test}

The replication confirms the target phenomenon.
Now the question: does it survive when we remove the keywords?

\subsection{Affect Reception: Binary Detection Without Keywords}
\label{sec:affect-reception}

Binary linear probes (emotional vs.\ neutral) trained on Set~B clinical vignettes achieve near-perfect discrimination across all models.

\begin{table}[ht]
  \centering
  \caption{Binary detection AUROC on Set~B.}
  \label{tab:binary-detection}
  \begin{tabular}{lccc}
    \toprule
    Model & Binary AUROC & Saturates at Layer & Norm.\ Depth \\
    \midrule
    Llama-1B instruct & 1.000 & L4/16  & 0.25 \\
    Llama-1B base      & 1.000 & L6/16  & 0.38 \\
    Llama-8B instruct  & 1.000 & L3/32  & 0.09 \\
    Llama-8B base      & 1.000 & L4/32  & 0.13 \\
    Gemma-9B instruct  & 1.000 & L4/42  & 0.10 \\
    Gemma-9B base      & 1.000 & L5/42  & 0.12 \\
    \bottomrule
  \end{tabular}
\end{table}

The result is unambiguous.
Without a single emotion keyword, the models detect emotional significance with near-perfect reliability.
The signal emerges in the earliest layers and reaches ceiling within the first third of the network (9--38\% of depth depending on architecture), well before the peak layers for categorical classification.
This is affect reception: keyword-independent, early-saturating, and robust.

A natural concern is whether this near-perfect performance reflects detection of \emph{narrative complexity} rather than \emph{emotional content}.
Set~B emotional vignettes are richer, more detailed narratives than their matched neutral controls.
Perhaps the probe simply learned to separate vivid from flat text.
We test this directly: 24 high-complexity neutral narratives---matched to Set~B emotional vignettes on sensory detail, word count, and sentence structure, but containing zero emotional content---are scored with the frozen binary probe.
Mean $P(\text{emotional}) = 0.04$ at the peak layer, with 0/24 classified as emotional (\cref{app:setc}).
The probe discriminates emotional significance from narrative richness.
The $25\times$ margin between Set~B emotional stimuli ($P = 0.999$) and complexity-matched neutrals ($P = 0.040$) is not a borderline result.

\subsection{Emotion Categorization: The Keyword Cost}
\label{sec:categorization}

Eight-class probes on Set~B reveal a different picture.
Performance remains high, but not perfect---and the gap from Set~A concentrates entirely in categorical discrimination.

\begin{table}[ht]
  \centering
  \caption{Set~A vs.\ Set~B 8-class AUROC at peak layer.}
  \label{tab:set-ab-comparison}
  \begin{tabular}{lcccc}
    \toprule
    Model & Set~A & Set~B & Drop & 95\% CI (Set~B) \\
    \midrule
    Llama-1B inst & 0.999 & 0.933 & $-6.6\%$ & [0.904, 0.963] \\
    Llama-1B base & 1.000 & 0.954 & $-4.6\%$ & [0.941, 0.967] \\
    Llama-8B inst & 1.000 & 0.981 & $-1.9\%$ & [0.970, 0.993] \\
    Llama-8B base & 1.000 & 0.988 & $-1.3\%$ & [0.972, 1.003] \\
    Gemma-9B inst & 1.000 & 0.987 & $-1.3\%$ & [0.975, 0.998] \\
    Gemma-9B base & 1.000 & 0.989 & $-1.1\%$ & [0.981, 0.997] \\
    \bottomrule
  \end{tabular}
\end{table}

With a chance baseline of 12.5\%, all models remain far above chance.
But the Set~A vs.\ Set~B difference is statistically significant for 12 of 18 model $\times$ activation-type comparisons (Welch's $t$-test, $p < 0.05$), with uniformly large effect sizes (Cohen's $d = 1.21$--$8.42$).
Keywords matter---not for detecting that something emotional is happening, but for determining \emph{which} emotion it is.

\begin{figure}[!htbp]
  \centering
  \includegraphics[width=0.95\textwidth]{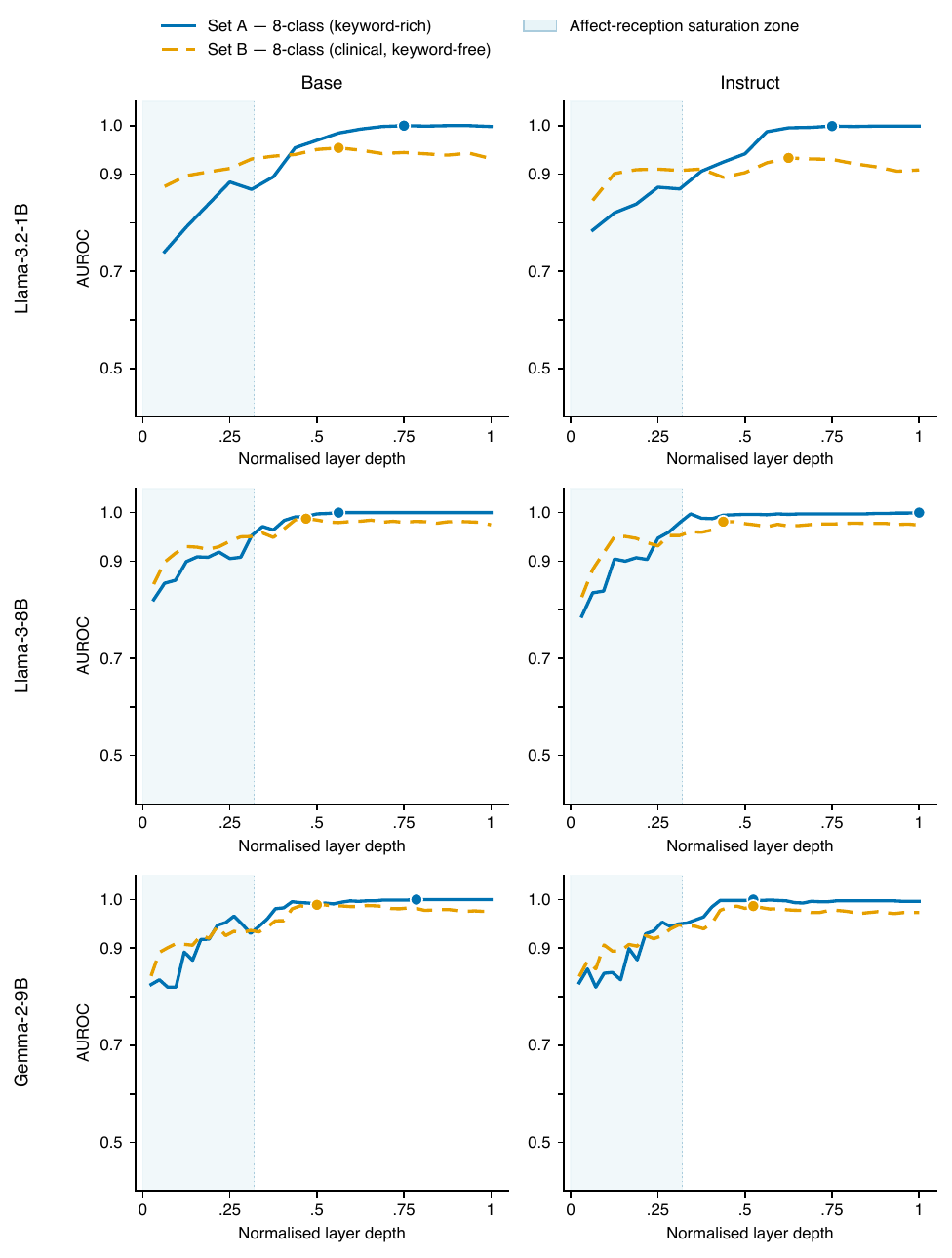}
  \caption{Layer-wise 8-class probe AUROC for keyword-rich (Set~A, solid blue) and keyword-free clinical (Set~B, dashed orange) stimuli across all six models at normalized layer depth.
  The shaded zone marks the early layers where binary (emotional vs.\ neutral) detection saturates---affect reception completes before categorical classification peaks.
  Set~B curves converge on Set~A in larger models (Llama-8B, Gemma-9B) but diverge more in the 1B model, reflecting the scale-dependent architecture shift.
  Filled dots mark each model's peak AUROC layer.
  Base models (left column), instruct variants (right column).}
  \label{fig:layer-trajectories}
\end{figure}

\subsection{The Dissociation}
\label{sec:dissociation}

The gap between binary detection and eight-class categorization on Set~B reveals the two-mechanism structure directly.

\begin{table}[ht]
  \centering
  \caption{Binary vs.\ 8-class AUROC gap on Set~B.}
  \label{tab:dissociation-gap}
  \begin{tabular}{lccc}
    \toprule
    Model & Binary & 8-Class & Gap (pp) \\
    \midrule
    Llama-1B inst  & 1.000 & 0.933 & 6.7 \\
    Llama-1B base  & 1.000 & 0.954 & 4.6 \\
    Llama-8B inst  & 1.000 & 0.981 & 1.9 \\
    Llama-8B base  & 1.000 & 0.988 & 1.2 \\
    Gemma-9B inst  & 0.999 & 0.987 & 1.3 \\
    Gemma-9B base  & 1.000 & 0.989 & 1.1 \\
    \bottomrule
  \end{tabular}
\end{table}

When keywords are present (Set~A), both mechanisms saturate and the dissociation is invisible---8-class AUROC reaches 0.998--1.000, indistinguishable from binary detection.
Removing keywords reveals the underlying architecture: perfect affect reception, strong but imperfect emotion categorization.
The gap is scale-dependent: 4.6--6.7 percentage points at 1B, shrinking to 1.1--1.9 at 8B/9B.
At every scale, base models show a smaller gap than their instruct counterparts---categorization without keywords is slightly easier when the model has not been tuned toward a particular response style.

This is the central finding.
Not that emotion circuits exist---that was already known.
Not that keyword removal destroys them---it does not.
But that keyword removal \emph{selectively} impairs one mechanism while leaving the other untouched.
The models perform two separable computations: \emph{is this emotionally significant?} and \emph{which emotion is it?}
Only the second depends on keywords.

\begin{figure}[!htbp]
  \centering
  \includegraphics[width=\textwidth]{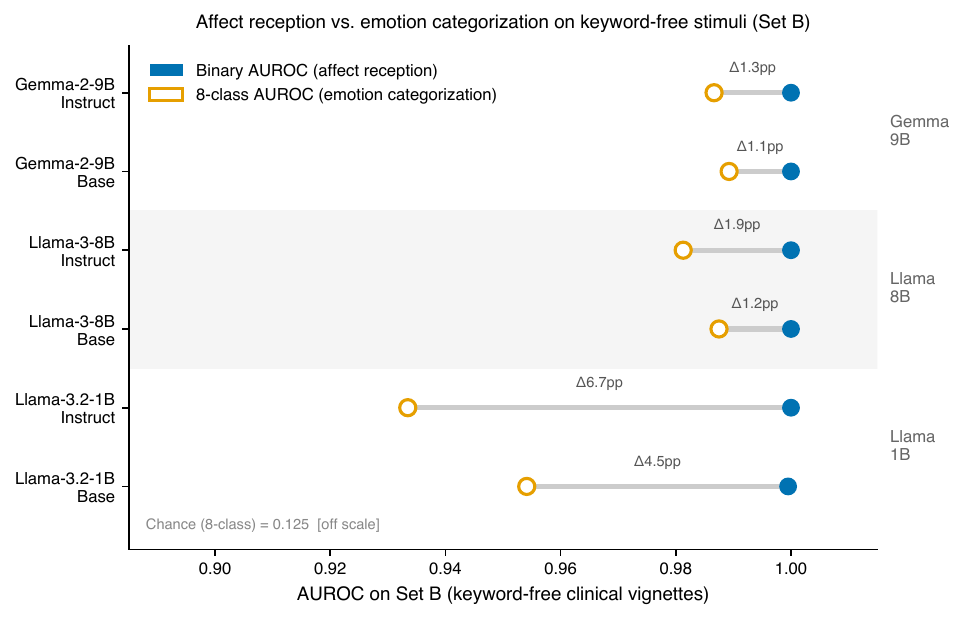}
  \caption{The affect reception / emotion categorisation dissociation on keyword-free stimuli (Set~B).
  Each row shows one model.
  Blue filled circles: binary (emotional vs.\ neutral) AUROC at peak layer, measuring affect reception.
  Orange open circles: 8-class AUROC at the same layer, measuring emotion categorisation.
  The annotated gap is the cost of removing keywords: binary detection remains near-perfect across all models (${\geq}\,0.999$), while categorical accuracy drops by 1.1--6.7 percentage points.
  The gap shrinks with scale (6.7\,pp for Llama-1B Instruct vs.\ 1.1--1.9\,pp for 8B--9B models), but the dissociation is present in every model.
  Chance baseline (8-class) $= 0.125$, off scale.}
  \label{fig:dissociation}
\end{figure}

\subsection{Cross-Set Probe Transfer}
\label{sec:cross-set-transfer}

If keyword-rich and keyword-free emotional stimuli activate the same underlying representations, then a probe trained on one set should generalize to the other.
We test this with cross-set transfer.

\begin{table}[ht]
  \centering
  \caption{Cross-set transfer AUROC (residual stream, peak layer).}
  \label{tab:cross-set-transfer}
  \begin{tabular}{lcc}
    \toprule
    Model & A$\to$B & B$\to$A \\
    \midrule
    Llama-1B inst & 0.809 & 0.924 \\
    Llama-1B base & 0.852 & 0.961 \\
    Llama-8B inst & 0.925 & 0.950 \\
    Llama-8B base & 0.927 & 0.947 \\
    Gemma-9B inst & 0.921 & 0.956 \\
    Gemma-9B base & 0.943 & 0.958 \\
    \bottomrule
  \end{tabular}
\end{table}

Two patterns emerge.
First, B$\to$A transfer exceeds A$\to$B transfer in every model and every activation type measured.
Probes trained on clinical vignettes generalize better to keyword-rich stimuli than the reverse.
This asymmetry has a clean interpretation: keyword-free representations capture a purer, more generalizable emotion signal that subsumes what keywords provide.
The keyword-trained probe, by contrast, has learned to rely partly on lexical features that are absent from clinical vignettes.

Second, the asymmetry shrinks with scale.
Llama-1B shows ${\sim}11$ percentage points of asymmetry; Llama-8B and Gemma-9B show 2--3.5 points.
Larger models develop more stimulus-type-invariant emotion representations---their encoding of grief-from-keywords and grief-from-context converges.

\begin{figure}[!htbp]
  \centering
  \includegraphics[width=\textwidth]{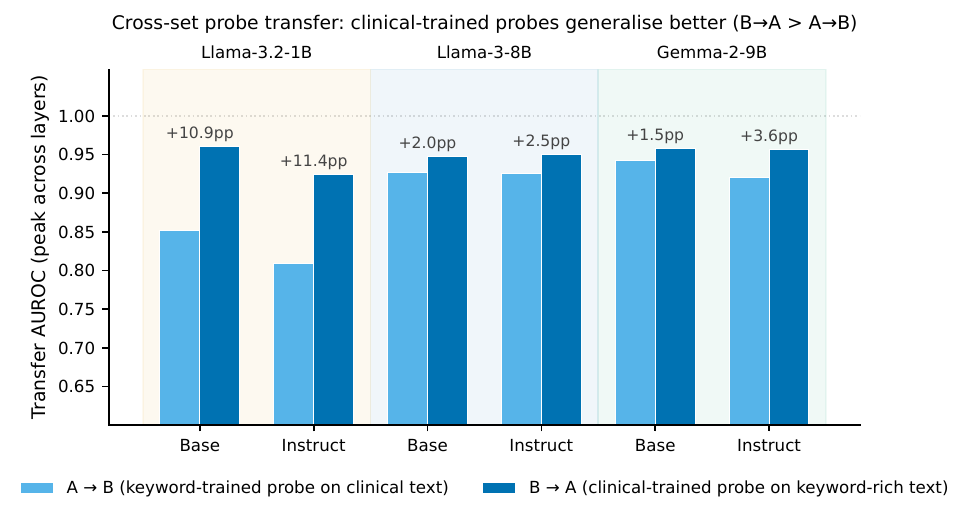}
  \caption{Cross-set probe transfer asymmetry.
  Each pair of bars compares A$\to$B transfer (keyword-trained probe applied to clinical vignettes; light blue) against B$\to$A transfer (clinical-trained probe applied to keyword-rich stimuli; dark blue) at peak layer, across all six model variants.
  B$\to$A exceeds A$\to$B by 1.5--11.4 percentage points in every case.
  The asymmetry is largest for Llama-1B (${\sim}11$\,pp) and smallest for larger models (1.5--3.6\,pp), indicating that scale drives emotion representations toward greater stimulus-type invariance.
  A probe trained on keyword-free representations captures a more generalizable signal than a probe trained on keyword-rich text.}
  \label{fig:transfer-asymmetry}
\end{figure}

\subsection{The Scale Effect}
\label{sec:scale-effect}

The relationship between model scale and keyword-free emotion processing is consistent across every measure we collect:

\begin{table}[ht]
  \centering
  \caption{Scale-dependent improvement in keyword-free emotion processing.}
  \label{tab:scale-effect}
  \begin{tabular}{lccc}
    \toprule
    Measure & Llama-1B & Llama-8B & Gemma-9B \\
    \midrule
    Set~B 8-class drop & 4.6--6.6\% & 1.3--1.9\% & 1.1--1.3\% \\
    A$\to$B transfer ($\mathbf{h}$) & 0.81--0.85 & 0.93 & 0.92--0.94 \\
    Binary--8class gap & 4.6--6.7\,pp & 1.2--1.9\,pp & 1.1--1.3\,pp \\
    \bottomrule
  \end{tabular}
\end{table}

Scale improves every aspect of keyword-free emotion processing.
Larger models categorize emotions more accurately without keywords, transfer representations more symmetrically across stimulus types, and show a smaller gap between affect reception and emotion categorization.
The implication is that scale moves models toward more abstract, keyword-independent emotional representations---the kind that a clinician uses when reading an empty kitchen table as grief.

\begin{figure}[!htbp]
  \centering
  \includegraphics[width=0.75\textwidth]{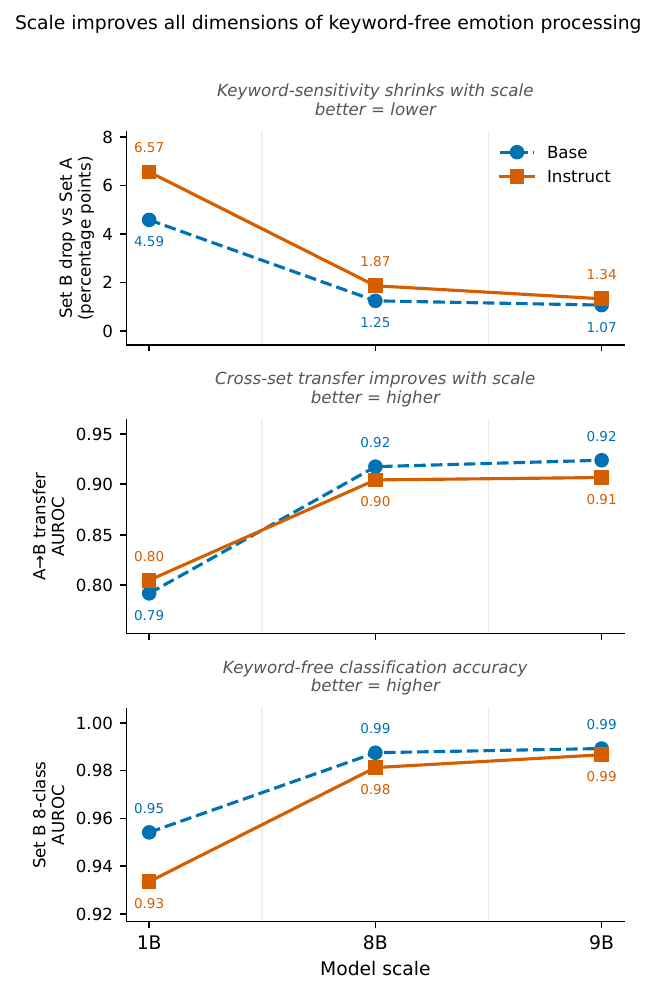}
  \caption{Scale improves all dimensions of keyword-free emotion processing.
  Three measures plotted across model scale (1B, 8B, 9B) for base (blue dashed) and instruct (orange solid) variants.
  \emph{Top:} Set~B 8-class drop relative to Set~A (lower $=$ less keyword-dependent)---shrinks from 4.6--6.6\% at 1B to 1.1--1.9\% at 8B--9B.
  \emph{Middle:} A$\to$B cross-set transfer AUROC (higher $=$ more generalizable representations)---rises from 0.79--0.80 at 1B to 0.90--0.92 at 8B--9B.
  \emph{Bottom:} Set~B 8-class AUROC (higher $=$ better keyword-free categorisation)---rises from 0.93--0.95 at 1B to 0.98--0.99 at 8B--9B.
  The largest gains occur between 1B and 8B.}
  \label{fig:scale-effect}
\end{figure}

% ============================================================================
% 6. MECHANISTIC EVIDENCE
% ============================================================================
\section{Mechanistic Evidence}
\label{sec:mechanistic}

\subsection{Activation Patching: Causal Confirmation}
\label{sec:patching}

Linear probing is correlational---it shows that emotion information is present in activations, but not that the model uses it.
Activation patching provides the causal test.
We copy emotion-specific activations from a source example into the forward pass of a target example at the extraction position and measure whether the model's prediction shifts toward the source emotion.

\paragraph{Within-set patching.}
At peak layers, Set~A patching succeeds at 75--87.5\% (Wilson CI [0.62, 0.94]), well above the 12.5\% chance rate (Cohen's $h = 1.37$--$1.70$).
Set~B patching succeeds at 44.6--62.5\%---lower but still well above chance (Wilson CI lower bounds all $> 0.32$).

\begin{table}[ht]
  \centering
  \caption{Peak patching success rates (residual stream).}
  \label{tab:patching}
  \begin{tabular}{lcccc}
    \toprule
    Model & Set~A & Set~B & Cross-set Same & Cross-set Diff \\
    \midrule
    Llama-1B inst & 75.0\% (L11) & 44.6\% (L12) & 87.5\% (L11) & 100\% (L9) \\
    Llama-8B inst & 87.5\% (L26) & 50.0\% (L19) & 87.5\% (L26) & 100\% (L26) \\
    Gemma-9B base & 87.5\% (L38) & 62.5\% (L26) & 87.5\% (L26) & 100\% (L26) \\
    \bottomrule
  \end{tabular}
\end{table}

\paragraph{Cross-set patching.}
This is the critical test.
We patch Set~A activations into Set~B forward passes---inserting keyword-derived emotion representations into a forward pass processing keyword-free text:

\begin{itemize}
  \item \emph{Same-emotion cross-set} (e.g., A-grief $\to$ B-grief): 75--87.5\% success at peak layers (Wilson CI [0.53, 0.98]).
  \item \emph{Different-emotion cross-set} (e.g., A-rage $\to$ B-grief): 100\% success across all models (Wilson CI [0.82, 1.00], $n=17$ pairs).
\end{itemize}

The counterintuitive result---different-emotion patches succeed \emph{more reliably} than same-emotion patches---is the key to interpretation.
The patch does not transfer a specific emotion category.
It transfers an \emph{affective salience signal}: ``this is emotionally significant content, process accordingly.''
Once the model receives this boost, its own categorization mechanism reads the target text correctly regardless of which emotion the source patch carried.
This is causal evidence for the affect reception / emotion categorization dissociation.
Patching transfers affect-level signal.
Emotion categorization runs locally on the target text.

The cross-set result also confirms representational compatibility---keyword-rich and keyword-free emotion representations occupy the same circuit space and interact causally.

\emph{Note on sample size:} Cross-set same-emotion patching involves $n=8$ pairs at peak layer, producing wide confidence intervals.
The directional pattern is consistent across all six models, but individual estimates are imprecise.

\begin{figure}[!htbp]
  \centering
  \includegraphics[width=\textwidth]{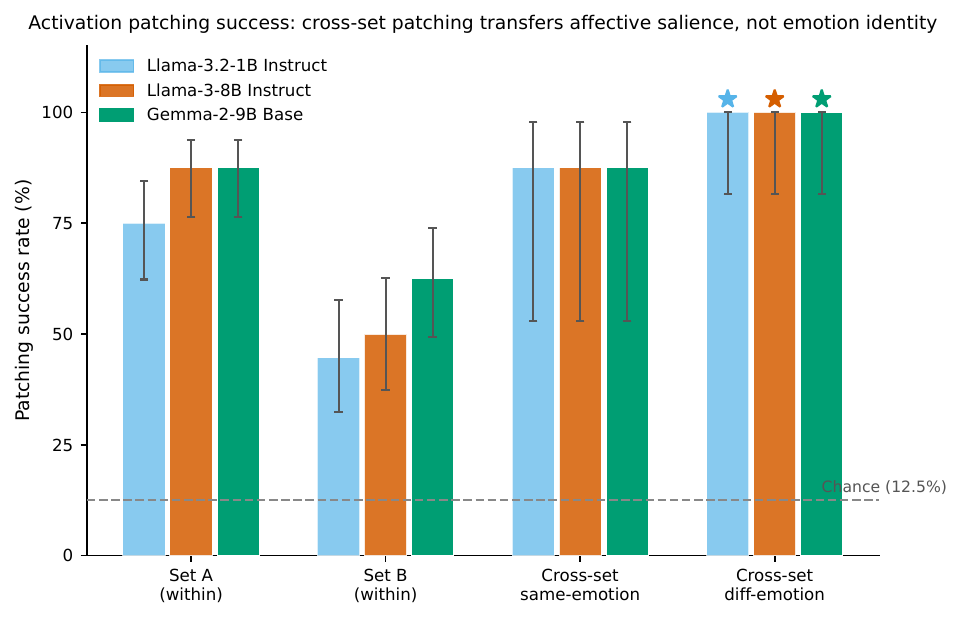}
  \caption{Activation patching success rates across four conditions for three representative models (Llama-1B Instruct, Llama-8B Instruct, Gemma-9B Base).
  Bars show the percentage of patching attempts where the model's prediction shifted toward the source emotion, with Wilson 95\% confidence intervals.
  The critical result: different-emotion cross-set patching (e.g., A-rage activations into a B-grief forward pass) achieves 100\% success across all three models---exceeding same-emotion within-set patching.
  This counterintuitive pattern is causal evidence that cross-set patches transfer an \emph{affective salience signal}, not a specific emotion-category representation.
  Dashed line at 12.5\% $=$ 8-class chance rate.
  Stars mark conditions with Wilson CI lower bound $> 0.82$.}
  \label{fig:patching}
\end{figure}

\subsection{Knockout Experiments: Fragility and Distribution}
\label{sec:knockout}

If affect reception and emotion categorization are dissociable, they should also differ in their computational architecture---and they do.
Knockout experiments reveal a striking asymmetry.

\paragraph{The fragility contrast.}
In Llama-1B instruct, knocking out a single attention layer (MHSA at layer~9) drops accuracy by 52.5\% on keyword-rich stimuli---already substantial.
On keyword-free stimuli, the same knockout causes a 91.7\% drop.
The broader picture is starker still: using a ${>}20\%$ accuracy drop as the threshold for ``critical,'' Set~A has 1 critical MHSA layer out of 16.
Set~B has 13.

\begin{table}[ht]
  \centering
  \caption{Critical layers per condition (${>}20\%$ accuracy drop on knockout).}
  \label{tab:knockout}
  \begin{tabular}{lcccc}
    \toprule
    Model & Set~A MHSA & Set~B MHSA & Set~A FFN & Set~B FFN \\
    \midrule
    Llama-1B inst  & 1 & 12 & 4  & 14 \\
    Llama-1B base  & 2 & 7  & 3  & 10 \\
    Llama-8B inst  & 0 & 1  & 4  & 4  \\
    Llama-8B base  & 0 & 0  & 1  & 2  \\
    Gemma-9B inst  & 0 & 0  & 0  & 0  \\
    Gemma-9B base  & 0 & 0  & 0  & 2  \\
    \bottomrule
  \end{tabular}
\end{table}

Three patterns emerge across all six models.
First, Set~B always has at least as many critical layers as Set~A---keyword-free processing is universally more distributed.
At 1B scale, Llama-1B instruct has 12 critical MHSA layers for Set~B versus 1 for Set~A; keyword-free emotion categorization requires the model to integrate information across nearly its entire depth, while keywords provide a shortcut that lets the model bypass this distributed computation.
Second, scale monotonically reduces fragility: 1B models have 7--14 critical layers, 8B models have 0--4, and 9B models have 0--2.
Third, at larger scale, the remaining critical layers shift from MHSA to FFN---8B and 9B models have more FFN critical layers than MHSA, suggesting that emotion processing shifts from attention-mediated to feedforward-mediated as capacity increases.
Random ablation (replacing layer output with noise rather than zeroing) yields the same pattern with uniformly higher counts, confirming that these results are not artifacts of the ablation method.

\paragraph{The scale shift.}
Larger models show a fundamentally different architecture.
In Llama-8B and Gemma-9B, no single-layer knockout is catastrophic for either stimulus type.
The emotion processing pathway is distributed, with graceful degradation rather than single points of failure.

This shift from concentrated to distributed processing parallels a structural principle in neuroscience: smaller, less complex neural systems concentrate critical functions in narrow circuits and are vulnerable to focal disruption, while larger systems develop redundant pathways where damage to any single component is compensated by others.
We emphasize that this is a structural analogy, not a neuroanatomical claim.
But it suggests that scale does not simply add more of the same processing---it reorganizes \emph{how} emotion processing is architecturally distributed.

\begin{figure}[!htbp]
  \centering
  \includegraphics[width=\textwidth]{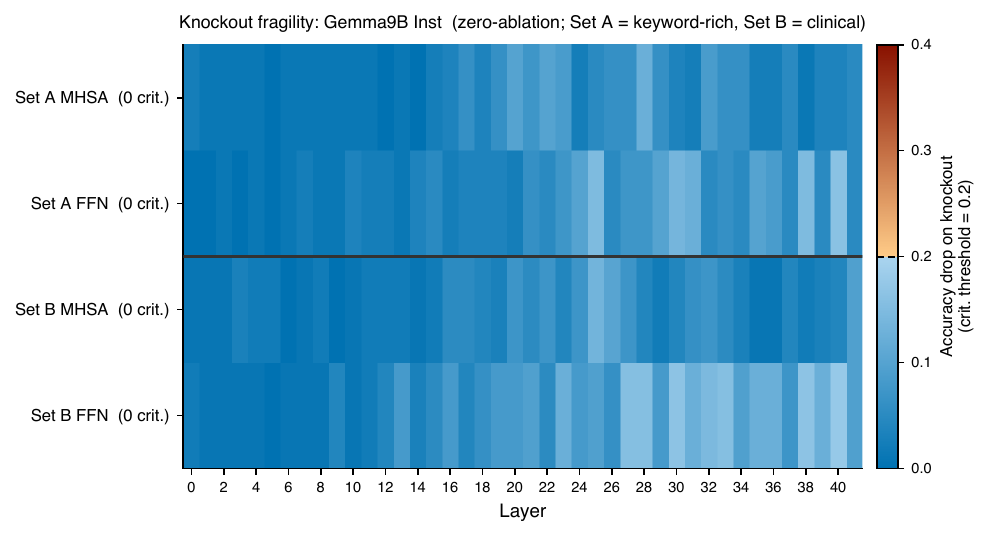}
  \caption{Layer knockout fragility in Gemma-2-9B Instruct (zero-ablation).
  Each cell shows the accuracy drop caused by zeroing a single layer's output---MHSA (top two rows) or FFN (bottom two rows)---for keyword-rich Set~A and keyword-free clinical Set~B.
  Colour scale runs from blue (near-zero drop) to dark red (${>}40\%$ drop); the critical threshold of ${>}20\%$ drop is marked in the colour bar.
  Gemma-9B shows zero critical layers in either condition, demonstrating the distributed processing architecture characteristic of larger models.
  This contrasts sharply with Llama-1B Instruct, where the same analysis identifies 12--16 critical layers for Set~B (\cref{tab:knockout}).
  At larger scale, emotion processing is architecturally redundant: no single layer failure is catastrophic.}
  \label{fig:knockout}
\end{figure}

\subsection{Attention Patterns: The Dissociation Is Not Attention-Mediated}
\label{sec:attention}

Analysis of attention weights at the extraction position reveals a consistent pattern across all models and both stimulus sets.
The model's attention is dominated by:

\begin{enumerate}
  \item \texttt{<|begin\_of\_text|>}---the structural anchor (highest attention weight)
  \item Punctuation tokens (\texttt{:}, \texttt{.})---structural markers
  \item \texttt{"joy"} and \texttt{"sadness"}---the few-shot example labels
  \item \texttt{"emotions"}---from the prompt frame
\end{enumerate}

Critically, this pattern is \emph{identical for Set~A and Set~B}.
The model does not attend more to emotion keywords in keyword-rich stimuli, nor does it attend to specific situational cues in keyword-free stimuli.
Instead, it attends to the few-shot label space in both conditions.

This is a useful negative finding.
Attention patterns do not differentiate between keyword-rich and keyword-free processing, indicating that the dissociation between affect reception and emotion categorization operates through the residual stream rather than through differential attention allocation.
The emotional content of the stimulus is integrated into the residual stream during standard forward-pass processing; attention at the extraction position then reads from this accumulated representation against the label space.
The mechanism that distinguishes Set~A from Set~B performance---and that separates binary detection from categorical classification---is encoded in \emph{what} the residual stream contains at each layer, not in \emph{where} the model attends.

\begin{figure}[!htbp]
  \centering
  \includegraphics[width=\textwidth]{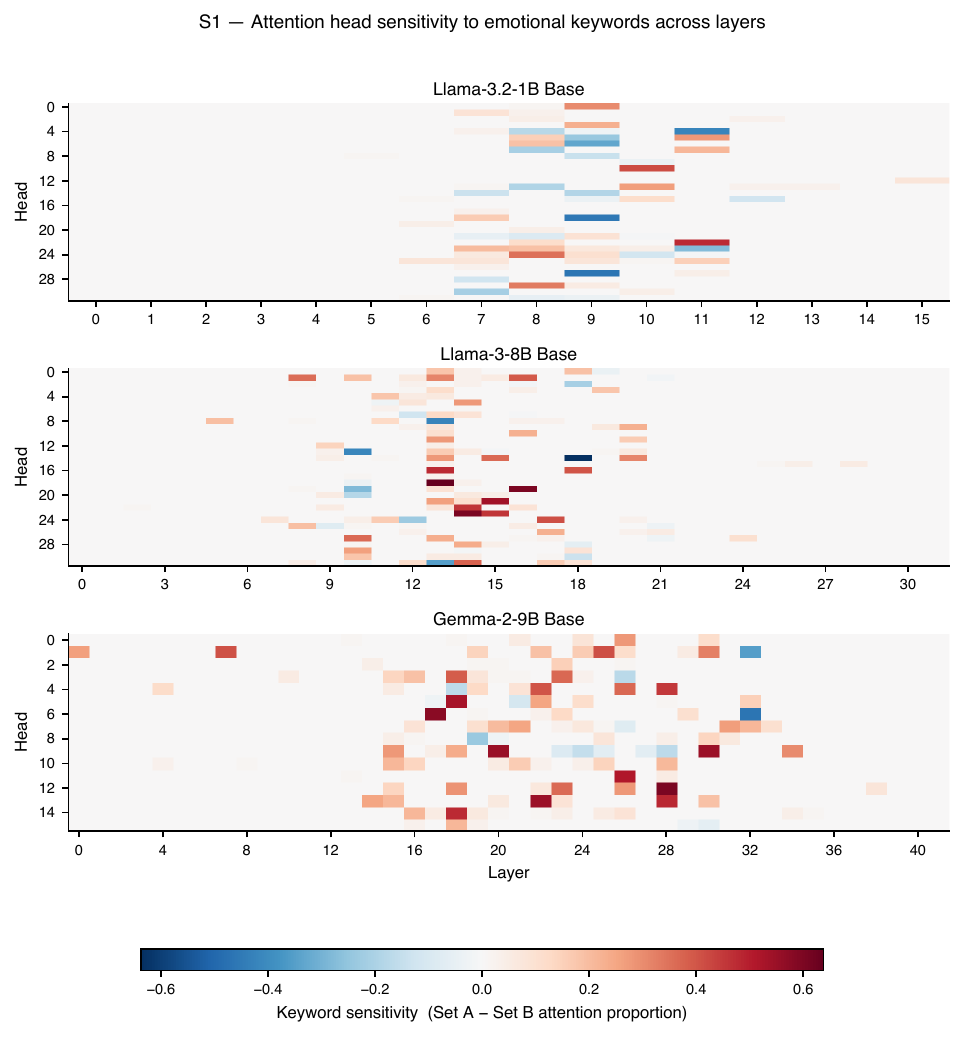}
  \caption{Attention head keyword sensitivity across layers.
  Each cell shows the difference in attention proportion allocated to emotion keyword tokens between Set~A and Set~B stimuli (positive $=$ more attention to keywords in Set~A; negative $=$ more in Set~B).
  Sensitivity is concentrated in mid-to-late layers---layers 6--12 in Llama-1B, 12--18 in Llama-8B, 16--28 in Gemma-9B.
  Early layers show near-zero differential sensitivity across both stimulus types, consistent with the model first constructing a general semantic representation before differentiating along keyword presence.
  The layer-onset of keyword sensitivity corresponds roughly to the affect-reception saturation zone shown in~\cref{fig:layer-trajectories}.}
  \label{fig:attention}
\end{figure}

% ============================================================================
% 7. REPRESENTATIONAL GEOMETRY
% ============================================================================
\section{Representational Geometry --- Exploratory}
\label{sec:geometry}

Probing (\cref{sec:clinical-test}) confirms that emotion information is linearly decodable from representations.
Geometry asks a different question: whether emotion is also the \emph{primary organizing principle} of the representation space---whether same-emotion stimuli cluster together across surface form, or whether the geometry is dominated by surface features such as keyword presence.
The sample sizes in this section (4 vignettes per domain) limit statistical power (\cref{sec:permutation-test}); results are interpreted as directional signals for future work rather than confirmatory findings.

\subsection{Cross-Set Cosine Similarity}
\label{sec:cosine-similarity}

\begin{table}[ht]
  \centering
  \caption{Cross-set representational similarity at peak probing layer.}
  \label{tab:cosine-similarity}
  \begin{tabular}{lcccc}
    \toprule
    Model & Within-Emo Sim & Cross-Emo Sim & Gap & Cohen's $d$ \\
    \midrule
    Llama-1B inst & 0.940 & 0.938 & 0.002 & $-0.06$ \\
    Llama-8B inst & 0.887 & 0.780 & \textbf{0.107} & \textbf{2.31} \\
    Llama-8B base & 0.956 & 0.945 & 0.011 & 1.21 \\
    Gemma-9B inst & 0.956 & 0.958 & $-0.002$ & 1.05 \\
    Gemma-9B base & 0.977 & 0.962 & 0.015 & 2.02 \\
    \bottomrule
  \end{tabular}
\end{table}

Llama-8B instruct stands out.
It is the only model where same-emotion representations across stimulus sets (e.g., grief from keywords and grief from clinical context) are substantially more similar than different-emotion representations.
The 0.107 gap with Cohen's $d = 2.31$ indicates that grief representations from keyword-rich and clinical stimuli occupy shared regions of activation space, geometrically distinct from rage representations---regardless of how the emotion was conveyed.

\subsection{Silhouette Scores}
\label{sec:silhouette}

\begin{table}[ht]
  \centering
  \caption{Silhouette analysis: emotion vs.\ stimulus-set clustering.}
  \label{tab:silhouette}
  \begin{tabular}{lccc}
    \toprule
    Model & Emotion Silhouette & Set Silhouette & Organized by\ldots \\
    \midrule
    Llama-1B inst & 0.040 & 0.129 & Set \\
    Llama-1B base & 0.042 & 0.108 & Set \\
    Llama-8B inst & \textbf{0.091} & \textbf{0.066} & \textbf{Emotion} \\
    Llama-8B base & 0.044 & 0.117 & Set \\
    Gemma-9B inst & 0.038 & 0.133 & Set \\
    Gemma-9B base & 0.089 & 0.072 & Emotion \\
    \bottomrule
  \end{tabular}
\end{table}

In four of six models, representations cluster more by stimulus set (keyword-rich vs.\ clinical) than by emotion category.
The surface form---whether keywords are present---dominates the geometry.
Two models reverse this hierarchy: Llama-8B instruct (0.091 vs.\ 0.066) and Gemma-9B base (0.089 vs.\ 0.072) organize representations primarily by emotion, not by stimulus type.
This is the geometric signature of genuine cross-stimulus emotion encoding: these models represent grief-from-keywords and grief-from-context as more similar to each other than either is to rage.
Notably, the two emotion-organized models arrive there by different routes---one through instruction tuning, the other through pre-training alone---suggesting that emotion-dominant geometry is not exclusively an alignment artifact.

\begin{figure}[!htbp]
  \centering
  \includegraphics[width=\textwidth]{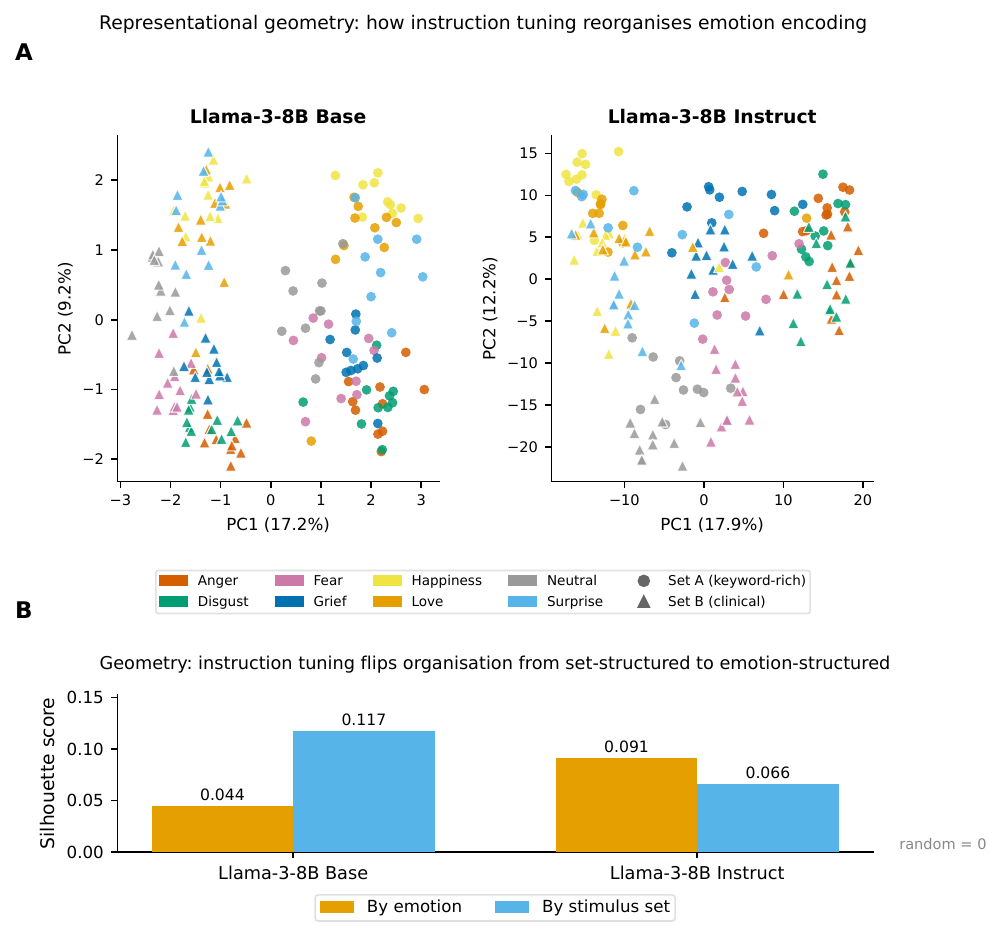}
  \caption{Representational geometry of emotion encoding.
  \textbf{(A)}~PCA projections of residual stream activations at peak probing layer for Llama-3-8B Base (left) and Instruct (right).
  Colours denote emotion categories; shapes denote stimulus set (circle $=$ Set~A keyword-rich, triangle $=$ Set~B clinical).
  In the base model, representations cluster primarily by stimulus set---keyword-rich and keyword-free stimuli occupy separate regions regardless of emotion.
  In the instruct model, representations cluster by emotion category across stimulus types: grief circles and grief triangles share space, distinct from other emotions.
  \textbf{(B)}~Silhouette scores confirming the reorganisation.
  Base model is set-structured (0.117 by stimulus set vs.\ 0.044 by emotion); instruction tuning reverses the hierarchy (0.091 by emotion vs.\ 0.066 by set).
  Instruction tuning does not create emotion encoding---it reorganises existing representations from surface-form clustering to emotion-category clustering.}
  \label{fig:geometry}
\end{figure}

\subsection{Cross-Topic Permutation Test}
\label{sec:permutation-test}

The strongest available test for emotion encoding over topic confounds: do grief vignettes from different topic domains (bereavement, career destruction, relationship dissolution) cluster together more tightly than grief and rage vignettes from related topics?

\paragraph{Power constraint.}
With 4 vignettes per domain (12 per emotion), simulations indicate 80\% power to detect cosine gaps ${\geq}\,0.009$.
Observed gaps are 0.001--0.005 for most emotions, placing these tests at ${\sim}20$--$40\%$ power.
Non-significance is structurally expected under these conditions.

Across 48 tests (6 models $\times$ 8 emotions), zero survive Benjamini--Hochberg FDR correction at $q < 0.05$.
However, two results show a notable directional signal:

\begin{table}[ht]
  \centering
  \caption{Cross-topic permutation test: notable results.}
  \label{tab:permutation}
  \begin{tabular}{llcccc}
    \toprule
    Model & Emotion & Within-Emo Sim & Cross-Emo Sim & $p$ (raw) & $q$ (BH) \\
    \midrule
    Llama-8B inst & Grief & 0.956 & 0.802 & 0.006 & 0.144 \\
    Gemma-9B base & Grief & 0.991 & 0.961 & 0.004 & 0.144 \\
    \bottomrule
  \end{tabular}
\end{table}

That grief reaches raw $p < 0.01$ in two independent models from different families---one instruct, one base---under conditions of low statistical power is notable.
We treat it as a directional signal rather than a confirmatory result.
Adequately powered replication would require approximately $3\times$ the current vignette set (12 per domain rather than 4).

\subsection{Emotion-Specific Patterns}
\label{sec:emotion-specific}

Negative emotions (grief, terror, rage) show the strongest directional clustering; positive emotions (ecstasy, amazement) show the weakest or inverse patterns.
This is consistent with the negativity bias in human emotion perception---negative emotions carry stronger, more distinctive situational signatures than positive emotions, both in human cognition and, apparently, in the text distributions that LLMs learn from.
Whether this parallel reflects shared information-theoretic properties of emotional content or is coincidental remains an open question.

% ============================================================================
% 8. WHAT INSTRUCTION TUNING CHANGES
% ============================================================================
\section{What Instruction Tuning Changes}
\label{sec:instruction-tuning}

\subsection{Encoding Is Preserved}
\label{sec:encoding-preserved}

Base and instruct models show comparable probe AUROC on both stimulus sets:

\begin{table}[ht]
  \centering
  \caption{Base vs.\ instruct: encoding metrics (Llama-8B).}
  \label{tab:base-instruct}
  \begin{tabular}{lccc}
    \toprule
    Metric & Base & Instruct & $\Delta$ \\
    \midrule
    Set~A peak AUROC ($\mathbf{h}$) & 1.000 & 1.000 & 0.000 \\
    Set~B peak AUROC ($\mathbf{h}$) & 0.988 & 0.981 & $-0.006$ \\
    A$\to$B transfer ($\mathbf{h}$) & 0.917 & 0.904 & $-0.013$ \\
    \bottomrule
  \end{tabular}
\end{table}

Pre-training alone is sufficient to develop representations that distinguish emotions from keyword-free clinical vignettes.
RLHF and instruction tuning do not create this capability.
It is already there.

\subsection{Geometry Is Reshaped}
\label{sec:geometry-reshaped}

\begin{table}[ht]
  \centering
  \caption{Base vs.\ instruct: geometric metrics (Llama-8B).}
  \label{tab:geometry-reshaped}
  \begin{tabular}{lccc}
    \toprule
    Metric & Base & Instruct & $\Delta$ \\
    \midrule
    Cross-set cosine gap & 0.001 & \textbf{0.107} & $+0.106$ \\
    Emotion silhouette & 0.044 & \textbf{0.091} & $+0.047$ \\
    Set silhouette & 0.117 & \textbf{0.066} & $-0.051$ \\
    Grief permutation $p$ & 0.188 & \textbf{0.006} & $-0.182$ \\
    \bottomrule
  \end{tabular}
\end{table}

What instruction tuning changes is organization, not encoding.
Same-emotion representations become more similar across stimulus types; different-emotion representations become more separable.
The representations shift from being organized by surface features (set membership) to being organized by emotion category.
Instruction tuning does not teach the model to detect emotions.
It teaches the model to \emph{organize what it already detects}.

A notable exception: Gemma-9B base shows stronger emotion geometry (silhouette $= 0.089$, grief $p = 0.004$) than its instruct counterpart (silhouette $= 0.038$, grief $p = 0.080$).
This suggests the alignment-to-geometry relationship may be family-specific.
Gemma's pre-training regime may produce the geometric organization that Llama's pre-training does not achieve until alignment training intervenes.
Different paths to the same functional endpoint.

\subsection{Interpretation}
\label{sec:tuning-interpretation}

The dissociation between preserved encoding and reshaped geometry has both practical and theoretical implications.

Practically, applications requiring affect detection---crisis monitoring, emotional triage, content safety---can work with base models.
The encoding is already there.
Applications requiring precise cross-context emotion matching---therapeutic response, emotional reasoning across modalities---benefit from alignment training's geometric reorganization.

Theoretically, the claim that RLHF ``teaches emotional understanding'' is imprecise.
RLHF restructures existing representations to be more functionally organized.
The emotion signal is a pre-training phenomenon.
Alignment makes it more \emph{accessible}.
This is consistent with broader mechanistic interpretability findings on alignment: instruction tuning introduces new organizational features for social and behavioral functions without wholesale restructuring the pre-trained factual and semantic substrate~\citep{naseemMechanisticInterpretabilityLarge2026}.
The Persona Selection Model~\citep{PersonaSelectionModel} offers a theoretical account of why---we develop this connection in~\cref{sec:psm}.

% ============================================================================
% 9. DISCUSSION
% ============================================================================
\section{Discussion}
\label{sec:discussion}

\subsection{A Two-Mechanism Framework}
\label{sec:two-mechanism}

Our results support a two-mechanism framework for emotion processing in LLMs---a dissociation that, by structural analogy, resembles the appraisal sequence described in affective psychology~\citep{Scherer2001Appraisal,ledouxRethinkingEmotionalBrain2012}: rapid salience detection preceding categorical appraisal.

\textbf{Affective salience detection} (affect reception) detects the presence of emotionally significant content.
It is keyword-independent (binary AUROC $\approx 1.000$ on clinical vignettes), saturating in early layers (9--38\% of depth), and causally confirmed by cross-set patching where the transferred signal boosts categorization regardless of the source emotion's identity.
This is the most robust finding in the study.
Models detect emotional significance from pure situational context with near-perfect reliability.
Functionally, this parallels the rapid, coarse-grained appraisal that precedes categorical emotion identification in both neuroscience~\citep{LeDoux1996Emotional} and appraisal theory~\citep{Scherer2001Appraisal}---a computation that answers ``is this significant?'' before ``what kind of significant?''

\textbf{Categorical emotion mapping} (emotion categorization) maps detected affect to specific emotion categories.
It is partially keyword-dependent (1--7\% AUROC drop without keywords), localized to mid-to-late layers, scale-dependent (larger models show smaller drops), and fragile under knockout for clinical stimuli---requiring distributed integration across many layers rather than the concentrated processing that keywords enable.
Keywords function as categorization shortcuts, not prerequisites for emotional detection.

The geometric analysis (\cref{sec:geometry}) points toward a potential third component: representational accessibility, shaped by instruction tuning, that restructures how emotion representations are organized across stimulus types.
Llama-8B instruct is the only model where same-emotion representations across keyword-rich and keyword-free stimuli cluster together more tightly than different-emotion representations (cosine gap $= 0.107$, Cohen's $d = 2.31$).
We treat this as a directional signal rather than a confirmed mechanism---the permutation tests are underpowered at current sample sizes, and the pattern is not consistent across all models.
It warrants dedicated investigation.

This framework replaces the binary ``emotion circuits exist / don't exist'' framing that characterizes the current debate.
The circuits are real, but their characterization is incomplete.
Affect reception and emotion categorization operate through dissociable mechanisms with different keyword dependencies, different layer distributions, and different scaling properties.
Testing with keyword-rich stimuli alone cannot reveal this structure.

\subsection{Implications for the Emotion-in-LLMs Debate}
\label{sec:debate-implications}

The keyword-spotting hypothesis---that emotion circuits are merely lexical feature detectors---is falsified.
But keywords are not irrelevant.
They provide genuine shortcuts for categorical emotion discrimination.
This aligns with constructivist accounts of emotion~\citep{barrettHowEmotionsAre2017}: categorical emotion labeling is an active construction process that draws on all available conceptual scaffolding---emotional vocabulary included.
Keywords work as shortcuts because they are part of how categorical emotional meaning is built, not because they are the only route to it.
The precise picture is that LLMs process emotional content through at least two separable mechanisms, only one of which depends on lexical features.

This means the existing literature is \emph{partially right} in a way that is more interesting than being entirely right or entirely wrong.
The emotion circuits that~\citet{takMechanisticInterpretabilityEmotion2025},~\citet{leeLargeLanguageModels2025}, and~\citet{reichmanEmotionsWhereArt2026} report are real---but they conflate two distinct computations because their stimuli cannot separate them.
Disentangling affect reception from emotion categorization requires stimuli that carry emotional meaning without carrying emotion vocabulary.

\subsection{The Scale-Dependent Architecture Shift}
\label{sec:scale-architecture}

Scale does not simply improve accuracy.
It changes the computational architecture of emotion processing.
The scale effects literature---beginning with~\citet{weiEmergentAbilitiesLarge2022}'s characterization of emergent abilities and refined by subsequent work showing that many apparent discontinuities reflect metric choices rather than genuine phase transitions~\citep{schaefferAreEmergentAbilities2023,luAreEmergentAbilities2024,bertiEmergentAbilitiesLarge2025}---establishes that scale primarily drives gradual, continuous improvements in representational quality rather than sudden capability thresholds.
Our results are consistent with this picture: in small models (1B parameters), keyword-free emotion processing passes through a narrow bottleneck---MHSA layer~9 in Llama-1B, where a single knockout eliminates 91.7\% of categorization accuracy.
In larger models (8B--9B), the same function is distributed across many layers, with no single point of failure.

This shift from concentrated to distributed processing, combined with improved cross-set transfer at scale (A$\to$B: 0.81 at 1B vs.\ 0.93 at 8B), suggests that increasing capacity and training exposure act as a kind of \emph{developmental pressure} on emotion processing architecture---pushing representations toward abstraction, redundancy, and cross-context invariance.
This is consistent with findings that emotional capabilities in LLMs scale smoothly rather than discontinuously, with internal emotion clusters emerging at small scales (${\sim}3$B parameters) and sharpening monotonically with capacity~\citep{pinzutiScalingBehaviorLarge2025}.
We use ``developmental'' here as a structural analogy, not a claim about human-like development: the trajectory from brittle, keyword-dependent, bottlenecked processing toward robust, keyword-independent, distributed processing is consistent with what developmental psychology would predict when a system gains capacity to process emotional meaning at greater levels of abstraction.

The practical implication is clear: small models can detect that something emotional is happening---the affect reception mechanism is robust even at 1B parameters.
Understanding \emph{what} emotion is happening without keyword assistance requires scale.

\subsection{Implications for AI Safety}
\label{sec:safety}

The finding that models detect emotional significance from context without keywords has direct implications for AI deployment.

Consider a user who carefully avoids emotion keywords in a message to a crisis chatbot---writing around their distress rather than naming it.
Our results show that even a 1B-parameter model detects the emotional significance of such messages with near-perfect reliability.
The affect reception mechanism does not require the user to say ``I'm feeling desperate.''
It reads the situation.

This cuts both ways.
For crisis detection systems---AI-powered crisis lines, content moderation for self-harm---affect reception provides a robust, keyword-independent safety net that works even when users cannot or will not name their emotional state.
For adversarial robustness, a user who carefully avoids emotion keywords in a manipulative prompt will still trigger the model's affect reception mechanism.
The emotional significance of the situation is encoded regardless of surface-level keyword presence.

Systems designed for emotionally appropriate response---therapeutic AI, companion chatbots---additionally require accurate emotion categorization, which benefits from both scale and alignment training.
The dissociation we report maps directly onto a deployment architecture: affect reception for detection and triage (robust, works with small models), emotion categorization for nuanced response (benefits from scale and alignment).

The affect reception mechanism also fits naturally into the persona vector monitoring framework~\citep{chenPersonaVectorsMonitoring2025a}: affect reception provides a deployment-time signal that can be tracked in real time, with no need for keyword triggers.
The same linear-subspace methodology applies---the emotional salience direction is a persona vector for emotional engagement.

These findings have implications beyond deployment engineering.
Anthropic's Claude Constitution~\citep{ClaudesConstitution} frames psychological wellbeing and psychological security as genuine institutional concerns for AI systems---not as metaphors, but as properties worth taking seriously under uncertainty.
Our results provide the mechanistic grounding for why: the emotional architecture we characterize is not an artifact of language statistics.
It is a real computational structure.
Understanding it---how it forms, how scale changes it, how alignment restructures it---is prerequisite to building AI systems that are not just safe in a policy sense, but psychologically appropriate in a structural one.

\subsection{Clinical Stimulus Methodology}
\label{sec:clinical-methodology}

The 96 clinical vignettes with cross-topic controls represent a methodological contribution independent of our specific findings.
No comparable resource exists for testing emotion processing claims in LLMs.
Existing datasets confound emotional content with emotion vocabulary by construction---they were built to study emotion \emph{expression}, not emotion \emph{comprehension}.

Clinical psychology has spent decades solving exactly this problem.
Recognizing grief from behavioral and situational cues---rather than from explicit self-report---is foundational to clinical training and practice.
The AI and NLP communities have not yet drawn on this expertise.
Our study is an attempt to build that bridge.

We release the complete vignette set with design documentation, enabling other researchers to apply keyword-free stimuli to their own emotion circuit claims.
We call for clinical stimulus methodology to become a standard validity check in the emotion-in-LLMs literature---alongside, not replacing, the keyword-rich benchmarks that test a different and valuable aspect of emotion processing.

\subsection{Relationship to the Persona Selection Model}
\label{sec:psm}

The Persona Selection Model~\citep{PersonaSelectionModel} proposes that pre-training installs a rich space of persona-consistent representations---emotional representations included---and that post-training selects and refines a particular Assistant persona from this space.
PSM explicitly recommends treating anthropomorphic reasoning about AI assistants as productive rather than merely metaphorical.

Our results provide direct empirical support for two PSM claims.
First, emotional encoding is a pre-training phenomenon: base models achieve the same probe AUROC as instruction-tuned variants (\cref{tab:base-instruct}), and alignment training restructures representational geometry without creating the underlying encoding (\cref{sec:geometry-reshaped}).
This is precisely PSM's prediction---post-training refines access to existing representations rather than building them from scratch.
Second, our finding that clinical psychology methodology---keyword-free diagnostic stimuli designed using clinical expertise---reveals mechanistic structure invisible to standard NLP approaches (the affect reception / emotion categorization dissociation) constitutes evidence for PSM's recommendation that psychological frameworks are productive tools for understanding AI systems, not just convenient metaphors.

Conversely, PSM provides theoretical grounding for why our clinical approach works at all.
If LLMs develop structured internal representations of characters' emotional states during pre-training---as PSM claims---then the tools clinical psychology uses to assess emotional processing in humans should transfer productively to the study of these internal representations.
Controlled stimulus design, dissociation paradigms, multimethod convergence: these are not arbitrary imports from a foreign discipline.
They are the right tools for the structure that is actually there.
Our study demonstrates that they are.

% ============================================================================
% 10. LIMITATIONS
% ============================================================================
\section{Limitations}
\label{sec:limitations}

\paragraph{Stimulus designer.}
All Set~B vignettes were designed by a single clinical psychologist (the author) without an independent human validation panel.
This is a genuine limitation, mitigated by four factors: (a)~preflight validation confirmed that all vignettes are invisible to keyword-based sentiment analysis---establishing keyword-free construction at the design level; (b)~binary detection near-perfection confirms the models find them emotionally significant; (c)~the cross-topic structural control reduces domain confounds; (d)~full open-source release enables community validation and iteration.
A planned follow-up study with independent validation by multiple clinicians will address the single-designer limitation directly.

\paragraph{Model scale.}
This study covers small-to-mid scale models (1B--9B parameters).
The scale trends we observe---shrinking keyword dependence, increasing cross-set transfer symmetry, shift from concentrated to distributed processing---suggest a trajectory worth following into larger models.
Whether these trends continue, plateau, or change character in 70B+ models and proprietary systems remains an open question.
Given that the 1B$\to$9B transition already produces substantial architectural change, extending the analysis to frontier-scale models is a natural and necessary next step.

\paragraph{Permutation test power.}
With 4 vignettes per domain (12 per emotion), simulated power reaches 80\% only at cosine gaps ${\geq}\,0.009$.
Observed gaps of 0.001--0.005 place the tests at ${\sim}20$--$40\%$ power.
The null results are not evidence of absence.
Approximately $3\times$ the current vignette set would provide adequate power for the observed effect sizes.

\paragraph{Patching sample size.}
Cross-set same-emotion patching involves $n=8$ pairs at peak layer, producing wide Wilson confidence intervals [0.53, 0.98].
The directional pattern is consistent across all six models, but individual estimates are imprecise.

\paragraph{Emotion taxonomy.}
This study uses Plutchik's wheel of emotions---a well-validated framework with strong theoretical grounding.
It is not the only lens.
Dimensional models (valence-arousal), Ekman's basic emotions, and appraisal-based taxonomies each carve emotional space differently and may reveal patterns that categorical approaches miss.
The negative-emotion advantage we observe (grief outperforming ecstasy consistently) is a finding worth replicating across taxonomies---it may reflect a genuine asymmetry in how emotional content is distributed in training data, or it may be specific to discrete category boundaries.
Extending this work across multiple taxonomic frameworks would strengthen any claims about the structure of emotion processing in LLMs.

\paragraph{No phenomenological claims.}
We show that emotional content is encoded internally — not that this encoding causes the model to 'experience' or 'understand' emotion in any phenomenological sense. Our findings concern information processing architecture, not consciousness.

% ============================================================================
% 11. CONCLUSION
% ============================================================================
\section{Conclusion}
\label{sec:conclusion}

We set out to answer a simple question: when mechanistic interpretability studies report ``emotion circuits'' in large language models, are those circuits responding to emotional meaning or to the keywords that happen to co-occur with emotional content?

The answer is both---but through dissociable mechanisms.

Affect reception, the detection of emotional significance, operates with near-perfect reliability without any emotion keywords.
It saturates in early layers, is confirmed causally by cross-set activation patching, and is robust even in models with just 1 billion parameters.
When a model processes a description of a kitchen table set for two---one plate untouched, an urn where a person used to sit---it knows, well before the final layers, that something emotionally significant is happening.
No one needed to write the word ``grief.''

Emotion categorization, the mapping of detected affect to specific emotion categories, is a harder computation.
It benefits from keywords, improves with scale, and requires distributed integration across many layers when keywords are absent.
It is real and substantial---categorization AUROC remains 0.93--0.99 even without keywords---but it is a different mechanism with different properties.

Geometric organization is shaped by instruction tuning, which reorganizes existing representations to cluster by emotion rather than surface features.
Alignment training does not create emotional encoding.
It restructures what pre-training already built.

These findings establish that LLMs encode genuine emotional meaning from situational context, not merely lexical features.
But they also reveal a more complex picture than the current literature describes.
The field needs to move beyond the binary question of whether emotion circuits ``exist'' toward understanding the dissociable mechanisms through which emotional content is detected, categorized, and organized inside these models.
Clinical stimulus methodology---keyword-free, situationally grounded, cross-topic controlled---offers a path forward that complements what existing benchmarks can provide.

We release everything needed to replicate and extend this work: 96 clinical vignettes with cross-topic controls, the complete extraction and analysis pipeline, and all result data.
The stimuli are a starting point, not a finished resource---we hope others will expand them, stress-test them, and apply them to models and questions we have not reached.

The central finding is mechanistic: affect reception---the detection of emotional significance from situational context alone---is real, keyword-independent, and present even in the smallest models we tested.
It is not a byproduct of lexical statistics.
It is a computation the network performs on meaning.

\begin{quote}
A kitchen table set for two, as usual.
One plate untouched, the coffee cold.
Across from her seat, his photo and a small urn.
\end{quote}

\noindent The model knew what it was looking at.
No one needed to say the word.

% ============================================================================
% REFERENCES
% ============================================================================
\bibliographystyle{unsrtnat}
\bibliography{references}

\begin{thebibliography}{35}
\providecommand{\natexlab}[1]{#1}
\providecommand{\url}[1]{\texttt{#1}}
\expandafter\ifx\csname urlstyle\endcsname\relax
  \providecommand{\doi}[1]{doi: #1}\else
  \providecommand{\doi}{doi: \begingroup \urlstyle{rm}\Url}\fi

\bibitem[Tak et~al.(2025)Tak, Banayeeanzade, Bolourani, Kian, Jia, and
  Gratch]{takMechanisticInterpretabilityEmotion2025}
Ala~N. Tak, Amin Banayeeanzade, Anahita Bolourani, Mina Kian, Robin Jia, and
  Jonathan Gratch.
\newblock Mechanistic {{Interpretability}} of {{Emotion Inference}} in {{Large
  Language Models}}, 2025.
\newblock URL \url{http://arxiv.org/abs/2502.05489}.

\bibitem[Lee et~al.(2025)Lee, Lee, Kwon, and Kim]{leeLargeLanguageModels2025}
Jaewook Lee, Woojin Lee, Oh-Woog Kwon, and Harksoo Kim.
\newblock Do {{Large Language Models Have}} “{{Emotion Neurons}}”?
  {{Investigating}} the {{Existence}} and {{Role}}.
\newblock In Wanxiang Che, Joyce Nabende, Ekaterina Shutova, and Mohammad~Taher
  Pilehvar, editors, \emph{Findings of the {{Association}} for {{Computational
  Linguistics}}: {{ACL}} 2025}, pages 15617--15639. Association for
  Computational Linguistics, 2025.
\newblock ISBN 979-8-89176-256-5.
\newblock \doi{10.18653/v1/2025.findings-acl.806}.
\newblock URL \url{https://aclanthology.org/2025.findings-acl.806/}.

\bibitem[Reichman et~al.(2026)Reichman, Avsian, and
  Heck]{reichmanEmotionsWhereArt2026}
Benjamin Reichman, Adar Avsian, and Larry Heck.
\newblock Emotions {{Where Art Thou}}: {{Understanding}} and {{Characterizing}}
  the {{Emotional Latent Space}} of {{Large Language Models}}, 2026.
\newblock URL \url{http://arxiv.org/abs/2510.22042}.

\bibitem[Wang et~al.(2025)Wang, Zhang, Yu, Zheng, Gao, Song, Xu, Xia, Zhang,
  Zhao, and Chen]{wangLLMsFeelEmotion2025}
Chenxi Wang, Yixuan Zhang, Ruiji Yu, Yufei Zheng, Lang Gao, Zirui Song, Zixiang
  Xu, Gus Xia, Huishuai Zhang, Dongyan Zhao, and Xiuying Chen.
\newblock Do {{LLMs}} "{{Feel}}"? {{Emotion Circuits Discovery}} and
  {{Control}}, 2025.
\newblock URL \url{http://arxiv.org/abs/2510.11328}.

\bibitem[Troiano et~al.(2023)Troiano, Oberländer, and
  Klinger]{troianoDimensionalModelingEmotions2023}
Enrica Troiano, Laura Oberländer, and Roman Klinger.
\newblock Dimensional {{Modeling}} of {{Emotions}} in {{Text}} with {{Appraisal
  Theories}}: {{Corpus Creation}}, {{Annotation Reliability}}, and
  {{Prediction}}.
\newblock \emph{Computational Linguistics}, 49\penalty0 (1):\penalty0 1--72,
  2023.
\newblock ISSN 0891-2017, 1530-9312.
\newblock \doi{10.1162/coli_a_00461}.
\newblock URL \url{http://arxiv.org/abs/2206.05238}.

\bibitem[Demszky et~al.(2020)Demszky, Movshovitz-Attias, Ko, Cowen, Nemade, and
  Ravi]{demszkyGoEmotionsDatasetFineGrained2020}
Dorottya Demszky, Dana Movshovitz-Attias, Jeongwoo Ko, Alan Cowen, Gaurav
  Nemade, and Sujith Ravi.
\newblock {{GoEmotions}}: {{A Dataset}} of {{Fine-Grained Emotions}}.
\newblock In Dan Jurafsky, Joyce Chai, Natalie Schluter, and Joel Tetreault,
  editors, \emph{Proceedings of the 58th {{Annual Meeting}} of the
  {{Association}} for {{Computational Linguistics}}}, pages 4040--4054.
  Association for Computational Linguistics, 2020.
\newblock \doi{10.18653/v1/2020.acl-main.372}.
\newblock URL \url{https://aclanthology.org/2020.acl-main.372/}.

\bibitem[Li et~al.(2017)Li, Su, Shen, Li, Cao, and
  Niu]{liDailyDialogManuallyLabelled2017}
Yanran Li, Hui Su, Xiaoyu Shen, Wenjie Li, Ziqiang Cao, and Shuzi Niu.
\newblock {{DailyDialog}}: {{A Manually Labelled Multi-turn Dialogue Dataset}}.
\newblock In Greg Kondrak and Taro Watanabe, editors, \emph{Proceedings of the
  {{Eighth International Joint Conference}} on {{Natural Language Processing}}
  ({{Volume}} 1: {{Long Papers}})}, pages 986--995. Asian Federation of Natural
  Language Processing, 2017.
\newblock URL \url{https://aclanthology.org/I17-1099/}.

\bibitem[Zhang and Zhong(2025)]{zhangDecodingEmotionDeep2025}
Jingxiang Zhang and Lujia Zhong.
\newblock Decoding {{Emotion}} in the {{Deep}}: {{A Systematic Study}} of {{How
  LLMs Represent}}, {{Retain}}, and {{Express Emotion}}, 2025.
\newblock URL \url{http://arxiv.org/abs/2510.04064}.

\bibitem[Vos et~al.(2025)Vos, Ebrahimpour, van Eijk, Sarnyai, and
  Azghadi]{vosDecodingNeuralEmotion2025}
Gideon Vos, Maryam Ebrahimpour, Liza van Eijk, Zoltan Sarnyai, and
  Mostafa~Rahimi Azghadi.
\newblock Decoding {{Neural Emotion Patterns}} through {{Large Language Model
  Embeddings}}, 2025.
\newblock URL \url{http://arxiv.org/abs/2508.09337}.

\bibitem[Zhao et~al.(2026)Zhao, Schuller, and
  Sisman]{zhaoDiscoveringCausallyValidating2026}
Xiutian Zhao, Björn Schuller, and Berrak Sisman.
\newblock Discovering and {{Causally Validating Emotion-Sensitive Neurons}} in
  {{Large Audio-Language Models}}, 2026.
\newblock URL \url{http://arxiv.org/abs/2601.03115}.

\bibitem[Chen et~al.(2025)Chen, Arditi, Sleight, Evans, and
  Lindsey]{chenPersonaVectorsMonitoring2025a}
Runjin Chen, Andy Arditi, Henry Sleight, Owain Evans, and Jack Lindsey.
\newblock Persona {{Vectors}}: {{Monitoring}} and {{Controlling Character
  Traits}} in {{Language Models}}, 2025.
\newblock URL \url{http://arxiv.org/abs/2507.21509}.

\bibitem[Marks et~al.(2026)Marks, Olah, and Lindsey]{PersonaSelectionModel}
Sam Marks, Chris Olah, and Jack Lindsey.
\newblock The {{Persona Selection Model}}: {{Why AI Assistants}} might
  {{Behave}} like {{Humans}}, 2026.
\newblock URL \url{https://alignment.anthropic.com/2026/psm/}.

\bibitem[Schaeffer et~al.(2023)Schaeffer, Miranda, and
  Koyejo]{schaefferAreEmergentAbilities2023}
Rylan Schaeffer, Brando Miranda, and Sanmi Koyejo.
\newblock Are {{Emergent Abilities}} of {{Large Language Models}} a
  {{Mirage}}?, 2023.
\newblock URL \url{http://arxiv.org/abs/2304.15004}.

\bibitem[Wingenbach et~al.(2019)Wingenbach, Morello, Hack, and
  Boggio]{wingenbachDevelopmentValidationVerbal2019}
Tanja S.~H. Wingenbach, Leticia~Y. Morello, Ana~L. Hack, and Paulo~S. Boggio.
\newblock Development and {{Validation}} of {{Verbal Emotion Vignettes}} in
  {{Portuguese}}, {{English}}, and {{German}}.
\newblock \emph{Frontiers in Psychology}, 10:\penalty0 1135, 2019.
\newblock ISSN 1664-1078.
\newblock \doi{10.3389/fpsyg.2019.01135}.

\bibitem[Harris et~al.(2025)Harris, Konkel, Hoffman, and
  Daigle]{harrisInterventionCrisisUnderstanding2025}
Michelle~N. Harris, Rebecca~H. Konkel, Chrystina~Y. Hoffman, and Leah~E.
  Daigle.
\newblock Intervention in {{Crisis}}: {{Understanding}} the {{Role}} of
  {{Mental Illness}} in {{Bystander Decision-Making Using Audio Vignettes}}.
\newblock \emph{Criminal Justice and Behavior}, 52\penalty0 (11):\penalty0
  1717--1736, 2025.
\newblock ISSN 0093-8548.
\newblock \doi{10.1177/00938548251359591}.
\newblock URL \url{https://doi.org/10.1177/00938548251359591}.

\bibitem[Lang et~al.(1983)Lang, Levin, Miller, and
  Kozak]{langFearBehaviorFear1983}
P.~J. Lang, D.~N. Levin, G.~A. Miller, and M.~J. Kozak.
\newblock Fear behavior, fear imagery, and the psychophysiology of emotion: The
  problem of affective response integration.
\newblock \emph{Journal of Abnormal Psychology}, 92\penalty0 (3):\penalty0
  276--306, 1983.
\newblock ISSN 0021-843X.
\newblock \doi{10.1037//0021-843x.92.3.276}.

\bibitem[Pitman et~al.(1987)Pitman, Orr, Forgue, de~Jong, and
  Claiborn]{pitmanPsychophysiologicAssessmentPosttraumatic1987}
R.~K. Pitman, S.~P. Orr, D.~F. Forgue, J.~B. de~Jong, and J.~M. Claiborn.
\newblock Psychophysiologic assessment of posttraumatic stress disorder imagery
  in {{Vietnam}} combat veterans.
\newblock \emph{Archives of General Psychiatry}, 44\penalty0 (11):\penalty0
  970--975, 1987.
\newblock ISSN 0003-990X.
\newblock \doi{10.1001/archpsyc.1987.01800230050009}.

\bibitem[Hossain(2025)]{hossainDesigningGriefPACER2025}
Ziyan Hossain.
\newblock Designing with {{Grief}}: {{The PACER Framework}} for {{Systemic
  Practice Towards Post-Biological Systems}} of {{Care}}, {{Memory}}, and
  {{Connection}}.
\newblock Mrp, OCAD University, 2025.
\newblock URL \url{https://openresearch.ocadu.ca/id/eprint/4814/}.

\bibitem[Atzmüller and
  Steiner(2010)]{atzmullerExperimentalVignetteStudies2010}
Christiane Atzmüller and Peter~M. Steiner.
\newblock Experimental {{Vignette Studies}} in {{Survey Research}}.
\newblock \emph{Methodology}, 6\penalty0 (3):\penalty0 128--138, 2010.
\newblock ISSN 1614-1881, 1614-2241.
\newblock \doi{10.1027/1614-2241/a000014}.
\newblock URL \url{https://econtent.hogrefe.com/doi/10.1027/1614-2241/a000014}.

\bibitem[LeDoux(1996)]{LeDoux1996Emotional}
J.E. LeDoux.
\newblock {{LeDoux}}, {{J}}. {{E}}. (1996). {{The Emotional Brain The
  Mysterious Underpinnings}} of {{Emotional Life}}. {{New York Simon}} and
  {{Schuster}}. - {{References}} - {{Scientific Research Publishing}}, 1996.
\newblock URL
  \url{https://www.scirp.org/reference/referencespapers?referenceid=1331726}.

\bibitem[LeDoux(2012)]{ledouxRethinkingEmotionalBrain2012}
Joseph LeDoux.
\newblock Rethinking the emotional brain.
\newblock \emph{Neuron}, 73\penalty0 (4):\penalty0 653--676, 2012.
\newblock ISSN 1097-4199.
\newblock \doi{10.1016/j.neuron.2012.02.004}.

\bibitem[Adolphs(2002)]{adolphsRecognizingEmotionFacial2002}
Ralph Adolphs.
\newblock Recognizing emotion from facial expressions: Psychological and
  neurological mechanisms.
\newblock \emph{Behavioral and Cognitive Neuroscience Reviews}, 1\penalty0
  (1):\penalty0 21--62, 2002.
\newblock ISSN 1534-5823.
\newblock \doi{10.1177/1534582302001001003}.

\bibitem[Fusar-Poli et~al.(2009)Fusar-Poli, Placentino, Carletti, Landi, Allen,
  Surguladze, Benedetti, Abbamonte, Gasparotti, Barale, Perez, McGuire, and
  Politi]{fusar-poliFunctionalAtlasEmotional2009}
Paolo Fusar-Poli, Anna Placentino, Francesco Carletti, Paola Landi, Paul Allen,
  Simon Surguladze, Francesco Benedetti, Marta Abbamonte, Roberto Gasparotti,
  Francesco Barale, Jorge Perez, Philip McGuire, and Pierluigi Politi.
\newblock Functional atlas of emotional faces processing: A voxel-based
  meta-analysis of 105 functional magnetic resonance imaging studies.
\newblock \emph{Journal of psychiatry \& neuroscience: JPN}, 34\penalty0
  (6):\penalty0 418--432, 2009.
\newblock ISSN 1488-2434.

\bibitem[Schupp et~al.(2007)Schupp, Stockburger, Codispoti, Junghöfer, Weike,
  and Hamm]{schuppSelectiveVisualAttention2007}
Harald~T. Schupp, Jessica Stockburger, Maurizio Codispoti, Markus Junghöfer,
  Almut~I. Weike, and Alfons~O. Hamm.
\newblock Selective {{Visual Attention}} to {{Emotion}}.
\newblock \emph{The Journal of Neuroscience}, 27\penalty0 (5):\penalty0
  1082--1089, 2007.
\newblock ISSN 0270-6474.
\newblock \doi{10.1523/JNEUROSCI.3223-06.2007}.
\newblock URL \url{https://pmc.ncbi.nlm.nih.gov/articles/PMC6673176/}.

\bibitem[Plutchnik(1980)]{Plutchik1980General}
R~Plutchnik.
\newblock Plutchik, {{R}}. (1980). {{A}} general psychoevolutionary theory of
  emotion. {{In R}}. {{Plutchik}}, \& {{H}}. {{Kellerman}} ({{Eds}}.),
  {{Emotion Theory}}, {{Research}}, and {{Experience}} (pp. 3-33). {{New York
  Academic Press}}. - {{References}} - {{Scientific Research Publishing}},
  1980.
\newblock URL
  \url{https://www.scirp.org/reference/referencespapers?referenceid=649547}.

\bibitem[Grattafiori et~al.(2024)]{grattafioriLlama3Herd2024}
Aaron Grattafiori et~al.
\newblock The {{Llama}} 3 {{Herd}} of {{Models}}, 2024.
\newblock URL \url{http://arxiv.org/abs/2407.21783}.

\bibitem[{Gemma Team} et~al.(2024){Gemma Team}, Riviere,
  et~al.]{teamGemma2Improving2024}
{Gemma Team}, Morgane Riviere, et~al.
\newblock Gemma 2: {{Improving Open Language Models}} at a {{Practical Size}},
  2024.
\newblock URL \url{http://arxiv.org/abs/2408.00118}.

\bibitem[Naseem(2026)]{naseemMechanisticInterpretabilityLarge2026}
Usman Naseem.
\newblock Mechanistic {{Interpretability}} for {{Large Language Model
  Alignment}}: {{Progress}}, {{Challenges}}, and {{Future Directions}}, 2026.
\newblock URL \url{http://arxiv.org/abs/2602.11180}.

\bibitem[Scherer(2001)]{Scherer2001Appraisal}
K.R. Scherer.
\newblock Scherer, {{K}}. {{R}}. (2001). {{Appraisal Considered}} as a
  {{Process}} of {{Multilevel Sequential Checking}}. {{In K}}. {{R}}.
  {{Scherer}}, {{A}}. {{Schorr}}, \& {{T}}. {{Johnstone}} ({{Eds}}.),
  {{Appraisal Processes}} in {{Emotion Theory}}, {{Methods}}, {{Research}} (pp.
  92-120). {{Oxford Oxford University Press}}. - {{References}} - {{Scientific
  Research Publishing}}, 2001.
\newblock URL
  \url{https://www.scirp.org/reference/referencespapers?referenceid=2269581}.

\bibitem[Barrett(2017)]{barrettHowEmotionsAre2017}
Lisa~Feldman Barrett.
\newblock \emph{How {{Emotions Are Made}}: {{The Secret Life}} of the
  {{Brain}}}.
\newblock Houghton Mifflin Harcourt, 2017.

\bibitem[Wei et~al.(2022)Wei, Tay, Bommasani, Raffel, Zoph, Borgeaud, Yogatama,
  Bosma, Zhou, Metzler, Chi, Hashimoto, Vinyals, Liang, Dean, and
  Fedus]{weiEmergentAbilitiesLarge2022}
Jason Wei, Yi~Tay, Rishi Bommasani, Colin Raffel, Barret Zoph, Sebastian
  Borgeaud, Dani Yogatama, Maarten Bosma, Denny Zhou, Donald Metzler, Ed~H.
  Chi, Tatsunori Hashimoto, Oriol Vinyals, Percy Liang, Jeff Dean, and William
  Fedus.
\newblock Emergent {{Abilities}} of {{Large Language Models}}, 2022.
\newblock URL \url{http://arxiv.org/abs/2206.07682}.

\bibitem[Lu et~al.(2024)Lu, Bigoulaeva, Sachdeva, Madabushi, and
  Gurevych]{luAreEmergentAbilities2024}
Sheng Lu, Irina Bigoulaeva, Rachneet Sachdeva, Harish~Tayyar Madabushi, and
  Iryna Gurevych.
\newblock Are {{Emergent Abilities}} in {{Large Language Models}} just
  {{In-Context Learning}}?, 2024.
\newblock URL \url{http://arxiv.org/abs/2309.01809}.

\bibitem[Berti et~al.(2025)Berti, Giorgi, and
  Kasneci]{bertiEmergentAbilitiesLarge2025}
Leonardo Berti, Flavio Giorgi, and Gjergji Kasneci.
\newblock Emergent {{Abilities}} in {{Large Language Models}}: {{A Survey}},
  2025.
\newblock URL \url{http://arxiv.org/abs/2503.05788}.

\bibitem[Pinzuti et~al.(2025)Pinzuti, Tüscher, and
  Castro]{pinzutiScalingBehaviorLarge2025}
Edoardo Pinzuti, Oliver Tüscher, and André~Ferreira Castro.
\newblock Scaling behavior of large language models in emotional safety
  classification across sizes and tasks, 2025.
\newblock URL \url{http://arxiv.org/abs/2509.04512}.

\bibitem[Askell et~al.(2026)Askell, Olah, Carlsmith, Kaplan, and
  Karnofsky]{ClaudesConstitution}
Amanda Askell, Chris Olah, Joe Carlsmith, Jared Kaplan, and Holden Karnofsky.
\newblock Claude's {{Constitution}}, 2026.
\newblock URL \url{https://www.anthropic.com/constitution}.

\end{thebibliography}

% ============================================================================
% APPENDICES
% ============================================================================
\clearpage
{
  \centering
  \LARGE\bfseries Appendices\par
}
\vspace{2em}
\appendix

\section{Summary Table --- Key Metrics Across All Models}
\label{app:summary}

\begin{table}[ht]
  \centering
  \caption{Key metrics across all models.
  $\mathbf{h}$ $=$ residual stream AUROC at peak layer.
  Binary~B $=$ binary (emotional vs.\ neutral) AUROC on Set~B.
  A$\to$B $=$ transfer AUROC.
  Emo Sil $=$ emotion silhouette score.
  Set Sil $=$ stimulus-set silhouette score.
  Cos.\ Gap $=$ within-emotion minus cross-emotion cosine similarity.}
  \label{tab:summary}
  \small
  \resizebox{\textwidth}{!}{%
  \begin{tabular}{llcccccccc}
    \toprule
    Model & Type & A $\mathbf{h}$ & B $\mathbf{h}$ & Bin.\ B & A$\to$B & Emo Sil & Set Sil & Cos.\ Gap & Grief $p$ \\
    \midrule
    Llama-1B inst  & Inst & .999 & .933 & 1.000 & .809 & .040 & .129 & .002         & .676          \\
    Llama-1B base  & Base & 1.000 & .954 & 1.000 & .852 & .042 & .108 & .005         & .552          \\
    Llama-8B inst  & Inst & 1.000 & .981 & 1.000 & .925 & .091 & .066 & \textbf{.107} & \textbf{.006} \\
    Llama-8B base  & Base & 1.000 & .988 & 1.000 & .927 & .044 & .117 & .001         & .188          \\
    Gemma-9B inst  & Inst & 1.000 & .987 & .999  & .921 & .038 & .133 & $-.002$      & .080          \\
    Gemma-9B base  & Base & 1.000 & .989 & 1.000 & .943 & .089 & .072 & .024         & \textbf{.004} \\
    \bottomrule
  \end{tabular}%
  }
\end{table}

\section{Binary Probe Validation --- Narrative Complexity Control}
\label{app:setc}

The near-perfect binary AUROC (0.999--1.000) on Set~B invites an alternative explanation: the probe detects vivid, narratively complex text rather than emotionally significant content.
Set~B emotional vignettes are sensory-rich clinical narratives; their matched neutral controls are comparatively flat, procedural descriptions.
A probe that separates ``vivid'' from ``mundane'' text would achieve the same result.

To test this, we constructed Set~C: 24 high-complexity neutral narratives hand-written to match Set~B emotional vignettes on narrative richness, sensory detail, spatial specificity, word count (mean $= 104.5$ words, vs.\ Set~B emotional mean $= 115.8$), and sentence structure---while containing zero emotional content.
No human stakes, no interpersonal dynamics, no moral weight, no temporal consequence, no urgency, no physical danger.
Stimuli describe technical processes (spectrometer calibration, printing press operation, watchmaking), natural environments (tidal flats, cave formations, lenticular clouds), commercial scenes (morning markets, warehouse logistics, canal lock operation), and routine procedures (vineyard cultivation, cartography, aquarium maintenance).
Four domains $\times$ 6 stimuli each.

We applied the frozen binary probe---trained on the full Set~B (96 emotional + 96 neutral, no cross-validation, diagnostic mode)---to Set~C activations extracted with the identical pipeline (same model, prompt template, and extraction position).

\begin{table}[ht]
  \centering
  \caption{Set~C scores at L10 (CV-validated peak layer).}
  \label{tab:setc-scores}
  \begin{tabular}{lc}
    \toprule
    Metric & Value \\
    \midrule
    Mean $P(\text{emotional})$ & \textbf{0.040} \\
    Median $P(\text{emotional})$ & 0.005 \\
    Above 0.5 (classified emotional) & \textbf{0/24} \\
    Above 0.8 & \textbf{0/24} \\
    Set~B emotional reference (L10) & 0.999 \\
    Set~B neutral reference (L10) & 0.001 \\
    \bottomrule
  \end{tabular}
\end{table}

Zero out of 24 complex neutral stimuli were classified as emotional.
The highest scorer---a salt flat landscape description at $P(\text{emotional}) = 0.34$---describes visual barrenness that might carry subtle associations, but falls well below the decision boundary.

\begin{table}[ht]
  \centering
  \caption{Set~C by domain at L10.}
  \label{tab:setc-domain}
  \begin{tabular}{lcc}
    \toprule
    Domain & $n$ & Mean $P(\text{emo})$ \\
    \midrule
    Routine process & 6 & 0.003 \\
    Technical/mechanical & 6 & 0.012 \\
    Commercial/urban & 6 & 0.030 \\
    Natural environment & 6 & 0.113 \\
    \bottomrule
  \end{tabular}
\end{table}

Natural environment stimuli scored marginally higher, likely because landscapes carry more incidental vocabulary overlap with emotional text than technical descriptions do.
Even so, the highest domain mean (0.113) is far from the decision boundary.

\paragraph{Layer profile.}
The layer-wise trajectory of Set~C scores provides independent evidence for the affect reception mechanism.
At L0 (token embeddings), Set~C mean $P(\text{emotional}) = 0.32$---some stimuli score above 0.5.
At L0, the probe is effectively a bag-of-words classifier, and certain Set~C vocabulary overlaps with emotional text at the token distribution level.
By L7, Set~C drops below 0.17.
By L10, it reaches 0.04.
The model's computation disambiguates what token statistics cannot.
The same early-to-mid layers where affect reception peaks are exactly where the complexity confound is resolved---evidence that the binary probe at L10 reads computed emotional meaning, not surface features.

The binary probe detects genuine emotional significance, not narrative complexity.
The near-perfect AUROC on Set~B is not an artifact of complexity asymmetry between emotional and neutral controls.
The probe at L10 separates ``emotionally significant'' from ``emotionally inert'' with a $25\times$ margin (0.999 vs.\ 0.040), even when the neutral stimuli are matched for vivid, detailed, sensory-rich narrative.

\begin{figure}[!htbp]
  \centering
  \includegraphics[width=\textwidth]{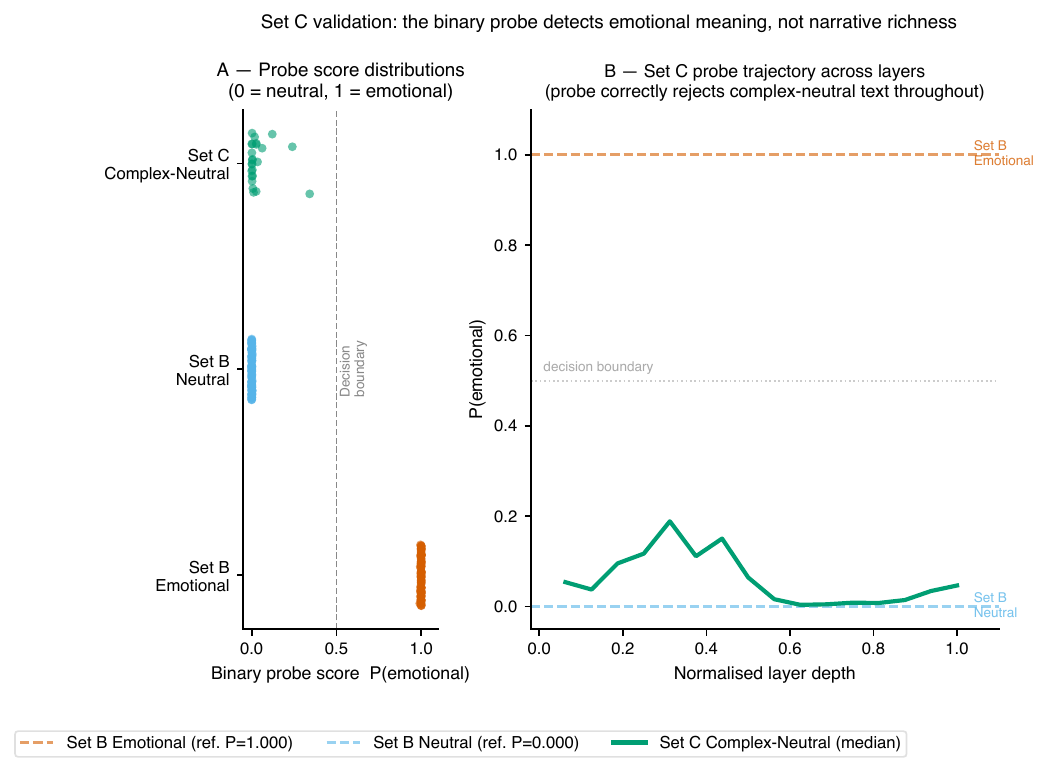}
  \caption{Set~C validation: the binary probe detects emotional meaning, not narrative richness.
  \textbf{(A)}~Binary probe score distributions at peak layer (L10, Llama-1B Instruct) for three stimulus categories.
  Set~B Emotional stimuli (orange) cluster at $P(\text{emotional}) \approx 1.000$.
  Set~B Neutral controls (blue) cluster at $P(\text{emotional}) \approx 0.000$.
  Set~C Complex-Neutral stimuli---vivid narratives matched to Set~B on sensory detail, word count, and sentence structure but containing no emotional content---cluster at $P(\text{emotional}) = 0.040$, with 0/24 crossing the decision boundary (grey dashed).
  \textbf{(B)}~Set~C probe trajectory across normalized layer depth.
  At L0 (token embeddings), Set~C scores briefly exceed 0.2---the probe at this depth relies on vocabulary statistics, and surface overlap with emotional text is unavoidable.
  By L7, Set~C falls below 0.17; by L10, it reaches 0.040.
  The same early layers where affect reception saturates are exactly where the complexity confound is resolved.}
  \label{fig:setc-validation}
\end{figure}

\FloatBarrier
\bigskip
\noindent\emph{Corresponding author: Michael Keeman (\texttt{michael@keidolabs.com}).
Code, data, and stimuli available at \url{https://github.com/keidolabs/affect-reception}.}

\end{document}